\documentclass[acmtog,authorversion, screen]{acmart}
\usepackage{booktabs} 
\usepackage{cuted}
\usepackage{xcolor, colortbl}

\acmJournal{TOG}
\acmYear{2020}
\acmVolume{39}
\acmNumber{4}
\acmArticle{75} 
\acmMonth{7}
\acmDOI{10.1145/3386569.3392457}

\setcopyright{acmlicensed}

\usepackage{hyperref}
\usepackage{url}

\definecolor{lightcyan}{rgb}{0.8,0.86,1}
\definecolor{darkgrey}{rgb}{0.5,0.5,0.5}
\usepackage{multirow}
\usepackage{overpic}

\newcommand{\mypercent}{\%}

\newcommand{\myreffig}[1]{Fig.~\ref{#1}}
\newcommand{\myreftab}[1]{Table~\ref{#1}}
\newcommand{\myrefsec}[1]{Sec.~\ref{#1}}
\newcommand{\myrefapp}[1]{Appendix~\ref{#1}}

\newcommand{\myavg}{\sum}

\newcommand{\resizeEq}[3]{\begin{equation}
\begin{aligned} #1 \label{#2}
\end{aligned}
\end{equation}}

\newcommand{\fnet}{$F$ }
\newcommand{\ppl}{PP}
\newcommand{\ofmae}{tOF}
\newcommand{\selflp}{tLP}
\newcommand{\TecoGANsmall}{{$\text{TecoGAN}^{\circleddash}$}}
\newcommand{\TecoSmallTwo}{{$^\text{Teco}_\text{GAN}{\circleddash}$}}
\newcommand{\selfpieapp}{tPieP}
\newcommand{\veryshortarrow}[1][3pt]{\mathrel{
\hbox{\rule[\dimexpr\fontdimen22\textfont2-.2pt\relax]{#1}{.4pt}}
\mkern-4mu\hbox{\usefont{U}{lasy}{m}{n}\symbol{41}}}}

\begin{document}
\title{Learning Temporal Coherence via Self-Supervision for GAN-based Video Generation}

\author{Mengyu Chu}
\authornote{Both authors contributed equally to the paper}
\email{mengyu.chu@tum.de}
\author{You Xie}
\authornotemark[1]
\email{you.xie@tum.de}
\author{Jonas Mayer}
\email{jonas.a.mayer@tum.de}
\author{Laura Leal-Taix\'{e}}
\email{leal.taixe@tum.de}
\author{Nils Thuerey}
\email{nils.thuerey@tum.de}

\affiliation{
\institution{\newline Technical University of Munich}
\department{Department of Computer Science}
\city{Munich}
\country{Germany}
}

\renewcommand\shortauthors{Chu, M.; Xie, Y.; Mayer, J.; Leal-Taix\'{e}, L.; Thuerey, N.}

\begin{abstract}
Our work explores temporal self-supervision for GAN-based video generation tasks.
While adversarial training successfully yields generative models
for a variety of areas, temporal relationships in the generated data
are much less explored.
Natural temporal changes are crucial for sequential generation tasks , e.g.
video super-resolution and unpaired video translation.
For the former, state-of-the-art methods often favor simpler norm losses such as $L^2$
over adversarial training. However, their averaging nature
easily leads to temporally smooth results with an undesirable lack of spatial detail.
For unpaired video translation, existing approaches modify the generator networks to
form spatio-temporal cycle consistencies.
In contrast, we focus on improving learning objectives and
propose a temporally self-supervised algorithm.
For both tasks, we show that temporal adversarial learning is key to achieving
temporally coherent solutions without sacrificing spatial detail.
We also propose a novel Ping-Pong loss to improve the long-term temporal consistency.
It effectively prevents recurrent networks from accumulating artifacts temporally
without depressing detailed features.
Additionally, we propose a first set of metrics to quantitatively evaluate the
accuracy as well as the perceptual quality of the temporal evolution.
A series of user studies confirm the rankings computed with these metrics.
Code, data, models, and results are provided at \url{https://github.com/thunil/TecoGAN}.
The project page \url{https://ge.in.tum.de/publications/2019-tecogan-chu/}
contains supplemental materials.
\end{abstract}




\begin{teaserfigure}
\centering
\includegraphics[width=1.0\textwidth]{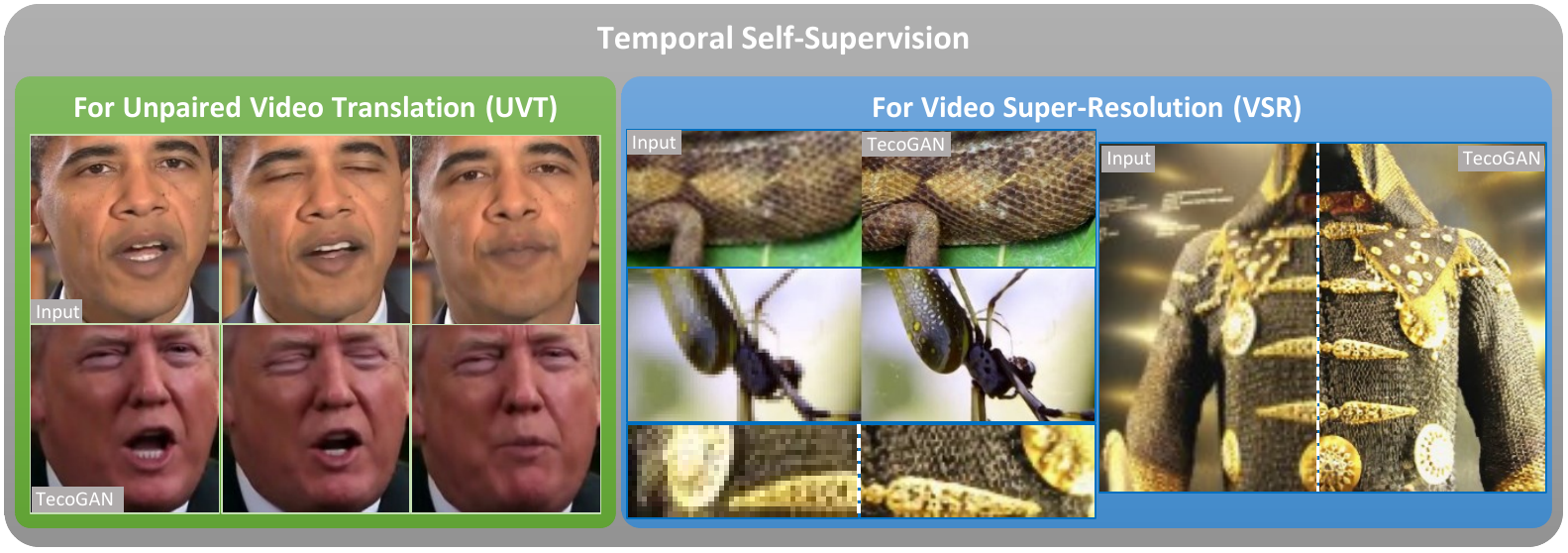}
\vspace{-12pt}
\caption{Using the proposed approach for temporal self-supervision, we achieve realistic results with natural temporal evolution for two inherently different video generation tasks: unpaired video translation (left) and video super-resolution (right).
While the resulting sharpness can be evaluated via the still images above,
the corresponding videos in \href{https://ge.in.tum.de/wp-content/uploads/2020/05/ClickMe.html}{our supplemental web-page}
{(Sec. 1 and Sec.2)} highlight the high quality of the temporal changes.
Obama and Trump video courtesy of the White House (public domain).}
\label{fig:teaser} \vspace{6pt}
\end{teaserfigure}

\maketitle

%%%%%%%%% BODY TEXT
\section{Introduction}

Generative adversarial networks (GANs)
have been extremely successful at 
learning complex distributions such as natural images~\cite{zhu2017cycleGAN, isola_2017_pix2pix}.
However, for sequence generation, directly applying {GANs}
without {carefully engineered constraints typically results in strong
artifacts over time due to the significant
difficulties introduced by the temporal changes.}
In particular, conditional video generation tasks are very challenging
learning problems where generators should
not only learn to represent the data distribution of the target domain but also learn to correlate the output distribution over time
with conditional inputs. Their central objective is to
faithfully reproduce the temporal dynamics of the target domain
and not resort to trivial solutions such as features that
arbitrarily appear and disappear over time.

In our work, we propose a novel adversarial learning method for a
recurrent training approach that supervises both spatial contents
as well as temporal relationships.
{As shown in \myreffig{fig:teaser}, }
we apply our approach to two video-related tasks that offer substantially different challenges: {\em video super-resolution} (VSR) and
{\em unpaired video translation} (UVT).
With no ground truth motion available, the spatio-temporal adversarial loss and
the recurrent structure enable our model to generate realistic results
while keeping the generated structures coherent over time.
With the two learning tasks we demonstrate how
spatio-temporal adversarial training 
can be employed in paired as well as
unpaired data domains.
In addition to the adversarial network which supervises the short-term temporal coherence,
long-term consistency is self-supervised using a novel bi-directional loss formulation, which we refer to as
``Ping-Pong'' (PP) loss in the following. The PP loss effectively avoids the temporal accumulation of artifacts,
which can potentially benefit a variety of recurrent architectures.
{We also note that most existing image metrics
focus on spatial content only. We fill the gap of
temporal assessment with a pair of metrics that measures the perceptual similarity over time and the similarity of motions with respect to a ground truth reference. User studies confirm these metrics for both tasks.}

The central contributions of our work are:
\begin{itemize}
\item {a spatio-temporal discriminator unit together with
a careful analysis of training} {objectives} 
for realistic and coherent video generation tasks, 
\item a novel PP loss supervising long-term consistency,
\item in addition to a set of metrics for quantifying temporal coherence based on motion estimation and perceptual distance.
\end{itemize}
Together, our contributions lead to models that outperform
previous work in terms of temporally-coherent detail, which we
{qualitatively and quantitatively demonstrate with a wide range of content}. 

\section{Related Work}
Deep learning has made great progress for image generation tasks.
While regular losses such as $L^2$~\cite{kim2016accurate, lai2017deep}
offer good performance for image super-resolution (SR) tasks in terms of PSNR metrics,
previous work found adversarial training~\cite{goodfellow2014generative}
to significantly improve the perceptual quality
in multi-modal settings such as
{
image generation~\cite{brock2018large},
colorization~\cite{He2018DeepExemplar},
}
super-resolution~\cite{ledig2016photo},
and translation~\cite{zhu2017cycleGAN, isola_2017_pix2pix}
tasks.
{Besides representing natural images,
GAN-based frameworks are also successful
at static graphic representations including}
{geometry synthesis~\cite{SAGnet19}
and city modeling~\cite{frankenGAN2018}.
}

{Sequential generation tasks, on the other hand, require
{the generation of realistic content that changes}
naturally over time~\cite{you2018tempoGAN, Kim2019smokestyle}.
It is especially so for conditional}
video generation tasks~\cite{StylizingVideo, HandheldMSR, EndToEndCam, zhang2019deep},
{where specific correlations between the input and the generated spatio-temporal evolution are required when ground-truth motions are not provided.
Hence, motion estimation~\cite{dosovitskiy2015flownet, liu2019selflow} and compensation become crucial for video generation tasks.
The compensation can take various forms, e.g., explicitly using variants of optical flow networks~\cite{shi2016real,caballero2017real,sajjadi2018FRVSR} and implicitly using deformable convolution layers~\cite{zhu2019deformable, wang2019edvr} or dynamic up-sampling~\cite{jo2018deep}.
In our work, a network is trained to estimate the motion and
we show that it can help generators and discriminators in spatio-temporal adversarial training.}

For VSR, recent work
improve the spatial detail and temporal coherence by either
using multiple low-resolution (LR) frames as inputs
\cite{haris2019recurrent, jo2018deep, tao2017spmc, liu2017robust}
or recurrently using previously estimated outputs
\cite{sajjadi2018FRVSR}.
The latter has the advantage to re-use high-frequency details over time.
In general, adversarial learning is less explored for VSR and applying it in conjunction with a recurrent structure gives rise
to a special form of temporal mode collapse, as we will explain below.
For video translation tasks, GANs are more commonly used but discriminators
typically only supervise the spatial content.
E.g., \citet{zhu2017cycleGAN} focuses on images without temporal constrains and generators
can fail to learn the temporal cycle-consistency for videos.
In order to learn temporal dynamics,
RecycleGAN~\cite{bansal2018recycle} proposes to use a
prediction network in addition to a generator,
while a concurrent work ~\cite{chen2019mocycle} chose to learn motion translation in addition to the spatial content translation.
Being orthogonal to these works, we propose a spatio-temporal adversarial training
for both VSR and UVT and we show that temporal self-supervision
is crucial for improving spatio-temporal correlations without sacrificing spatial detail.

While $L^1$ and $L^2$ temporal losses based on warping are generally used to enforce temporal smoothness in video style transfer tasks ~\cite{ruder2016artistic, chen2017coherent} and concurrent GAN-based VSR~\cite{perez2018photorealistic} and UVT~\cite{park2019preserving} work,
it leads to an undesirable smooth over spatial detail and temporal changes in outputs.
Likewise, the $L^2$ temporal metric represents a sub-optimal way to
quantify temporal coherence. {
For image similarity evaluation,
perceptual metrics~\cite{zhang2018unreasonable, prashnani2018pieapp} are proposed to reliably consider semantic features instead of pixel-wise errors. However, for videos, }
perceptual metrics that evaluate natural temporal changes are unavailable up to now.
To address this open issue, we propose two improved temporal metrics and demonstrate the advantages of
temporal self-supervision over direct temporal losses.
{
Due to its complexity, VSR has also led to workshop challenges like {\em NTIRE19}~\cite{Nah_2019_CVPR_Workshops},
where algorithms such as EDVR~\cite{wang2019edvr}
perform best w.r.t. PSNR-based metrics.
We compare to these methods and give additional details in \myrefapp{app:metrics}.}

Previous work, e.g. tempoGAN for fluid flow~\cite{you2018tempoGAN} and vid2vid for video translation~\cite{wang2018vid2vid}, has
proposed adversarial temporal losses to achieve time consistency.
While tempoGAN employs a second temporal discriminator with multiple aligned frames to assess the realism of temporal changes,
it is 
{not suitable for} videos, as it relies on ground truth motions and employs single-frame processing that is sub-optimal for natural images.
On the other hand, vid2vid focuses on paired video translations and proposes a video discriminator based on a conditional motion input that is estimated from the paired ground-truth sequences.
We focus on more difficult unpaired translation tasks instead and demonstrate the gains in the quality of our approach in the evaluation section.
{Bashkirova et al.~\shortcite{bashkirova2018unsupervised} solve UVT tasks as a 3D extension of the 2D image translation. In DeepFovea~\citep{kaplanyan2019deepfovea}, a 3D discriminator is used to supervise video in-painting results with 32 frames as a single 3D input. {Since temporal evolution differs from a spatial distribution, we show how a separate handling of the temporal dimension can reduce computational costs, remove
training restrictions, and most importantly improve  inference quality.}}

For tracking and optical flow estimation,
$L^2$-based time-cycle losses~\citep{wang2019learning} were proposed to constrain motions and tracked correspondences
using symmetric video inputs.
By optimizing indirectly via motion compensation or tracking,
this loss improves the accuracy of the results.
For video generation, we propose a PP loss that also makes use of symmetric sequences.
However, we directly constrain the PP loss via the generated video content,
which successfully improves the long-term temporal consistency in the video results.
{{The PP loss is effective by offering
valid information in forward as well as backward passes of image sequences.
This concept is also used in robotic control algorithms,}
where reversed trajectories starting from goal positions
have been used as training data~\citep{nair2018time}.}

\section{Learning Temporally Coherent Conditional Video Generation}
\label{sec:method}

{
We first propose the concepts of temporal self-supervision for
GAN-based video generation (\myrefsec{sec:dst} and \myrefsec{sec:pploss}), before
introducing solutions for VSR and UVT tasks
(\myrefsec{sec:VSR} and \myrefsec{sec:UVT})
as example applications.
}

\subsection{Spatio-Temporal Adversarial Learning}
\label{sec:dst}

\begin{figure}[bth]
\centering
\includegraphics[width=\linewidth]{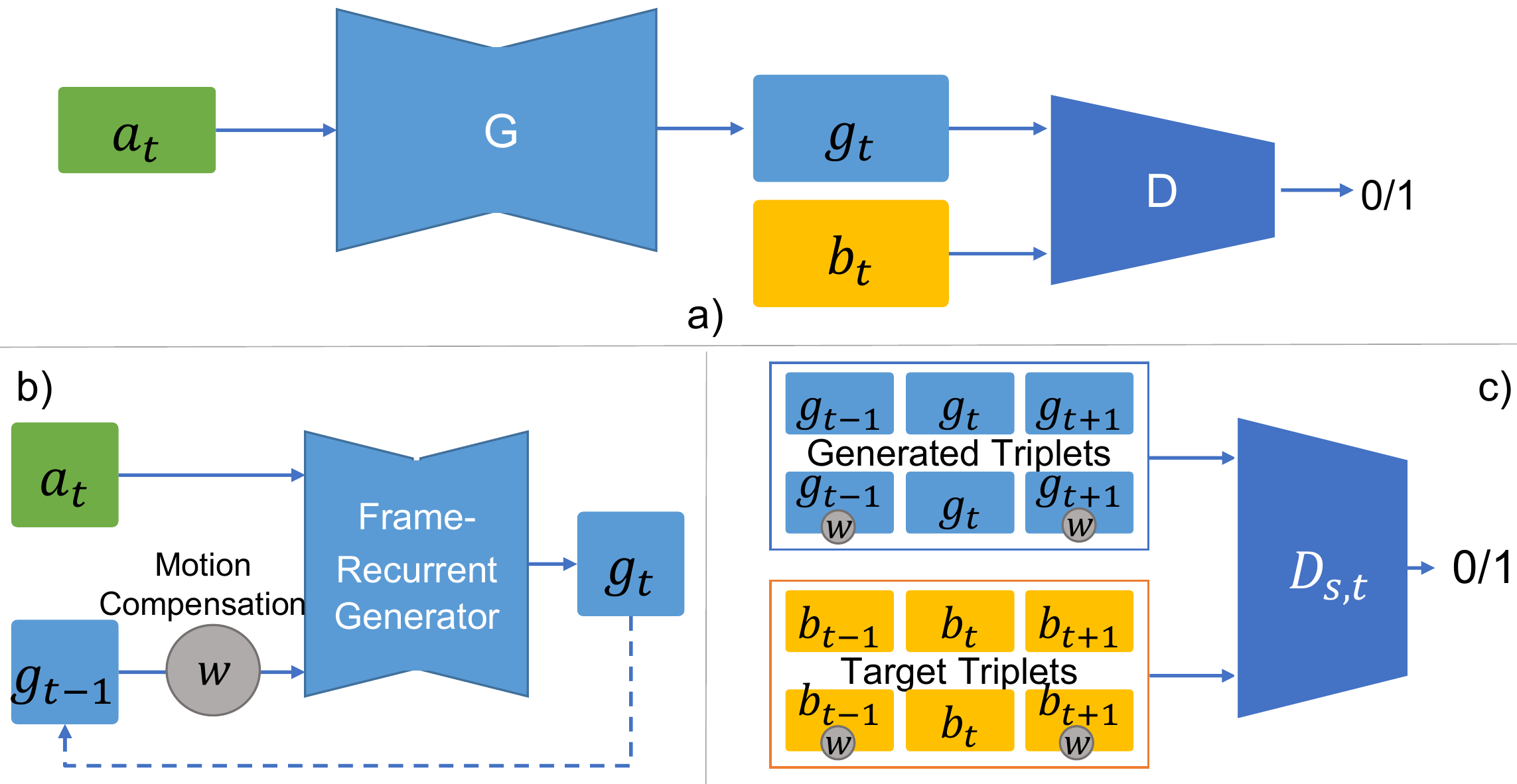}
\caption{{a) A spatial GAN for image generation. \quad
b) A frame recurrent Generator. \quad
c) A spatio-temporal Discriminator.
\quad In these figures, letter $a$, $b$, and $g$, stand for the input domain, the output domain and
the generated results respectively. $G$ and $D$ stand for the generator and the discriminator.}}
\label{fig:dst}
\end{figure}

{
\noindent
While GANs are popular and widely used in image generation tasks to improve perceptual quality,
their spatial adversarial learning {inherently} introduces temporal problems for tasks such as video generation. Thus, we propose an algorithm for spatio-temporal adversarial learning
that is easy to integrate into existing GAN-based image generation approaches.
Starting from a standard GAN for images, as shown in \myreffig{fig:dst} a), we propose
to use a frame-recurrent generator (b) together with a spatio-temporal discriminator (c).
}

{
As shown in \myreffig{fig:dst} b), our generator produces an output $g_t$ from
an input frame $a_t$ and recursively uses the previously generated output $g_{t-1}$.
Following previous work~\cite{sajjadi2018FRVSR}, we warp this frame-recurrent input
to align it with the current frame.
This allows the network to more easily re-use previously generated details.
The high-level structure of the generator can be summarized as:
\resizeEq{
v_t = \text{F}(a_{t-1}, a_t) , \quad
g_t = \text{G}( a_t,  W( g_{t-1}, v_t ) ) .
}{eq:generation}{0.75}
Here, the network \fnet is trained to estimate the motion $v_t$
from frame $a_{t-1}$ to $a_t$ and $W$ denotes warping.
}

The central building block of our approach is a novel {\em spatio-temporal}
discriminator $D_{s,t}$ that receives triplets of frames, shown in \myreffig{fig:dst} c).
This contrasts with typically used {\em spatial} discriminators that supervise only a single image.
By concatenating multiple adjacent frames along the channel dimension, the frame triplets form an important building block for
learning as they
can provide networks with gradient information regarding the realism of
spatial structures as well as short-term temporal information,
such  as first- and second-order time derivatives.

We propose a $D_{s,t}$ architecture
that primarily receives two types of triplets:
three adjacent frames and the corresponding warped ones.
We warp later frames backward and previous ones forward.
The network $F$ is likewise used to estimate the corresponding motions.
While original frames contain the full spatio-temporal information,
warped frames more easily yield temporal information with their aligned content.
For the input variants we use the following notations:
$\text{I}_g=\{g_{t-1}, g_t, g_{t+1}\}, \text{I}_b = \{b_{t-1}, b_t, b_{t+1}\}$;
$\text{I}_{wg} =\{ W( g_{t-1}, v_t ), g_{t}, W( g_{t+1}, v'_t )\}$,
$\text{I}_{wb} =\{ W( b_{t-1}, v_t ), b_{t}, W( b_{t+1}, v'_t ) \}$.
{A subscript $a$ denotes the input domain, while the $b$ subscript denotes the target domain.}
{The quotation mark in $v'$ indicates that quantities are estimated from the backward direction.}

{Although the proposed concatenation of several frames
seems like a simple change that has been used in a variety of {other contexts,
we show that it represents} an important operation that allows discriminators to understand
spatio-temporal data distributions.
As will be shown below, it can effectively reduce temporal problems encountered by spatial GANs.}
While $L^2-$based temporal losses are widely used in the field of video generation,
the spatio-temporal adversarial loss is crucial for preventing the inference of
blurred structures in multi-modal data-sets.
Compared to GANs using multiple discriminators,
the single $D_{s,t}$ network {that we propose} can learn to
balance the spatial and temporal aspects according to the reference data and
avoid inconsistent sharpness as well as overly smooth results.
Additionally, by extracting shared spatio-temporal features, it allows
for smaller network sizes.

\subsection{Self-Supervision for Long-term Temporal Consistency}
\label{sec:pploss}

\begin{figure}[t]
\begin{minipage}{0.34\linewidth}
\begin{overpic}[width=\linewidth]{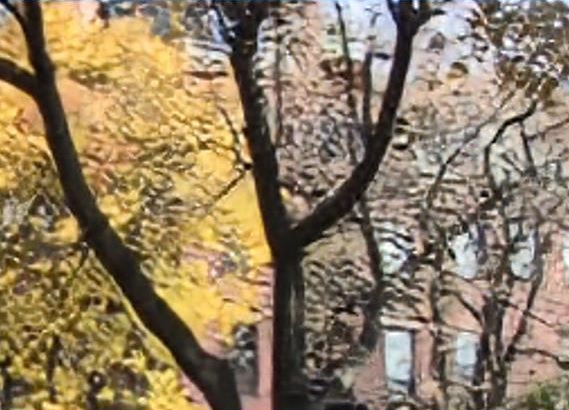}
\put(0,4){{ \color{white}{\textbf{a)}}}}
\end{overpic}\\\vspace{-8pt}
\begin{overpic}[width=\linewidth]{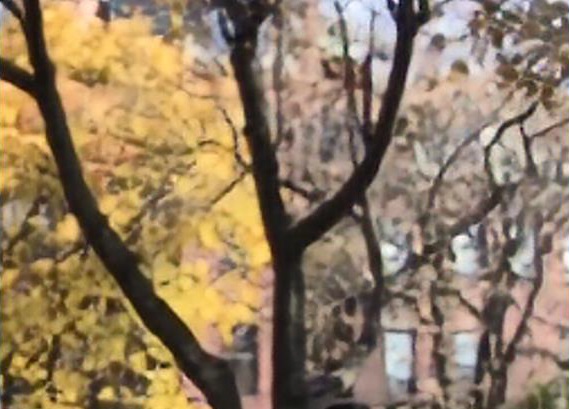}
\put(0,4){{ \color{white}{\textbf{b)}}}}
\end{overpic}
\end{minipage}
\begin{minipage}{0.65\linewidth}
\begin{overpic}[width=\linewidth]{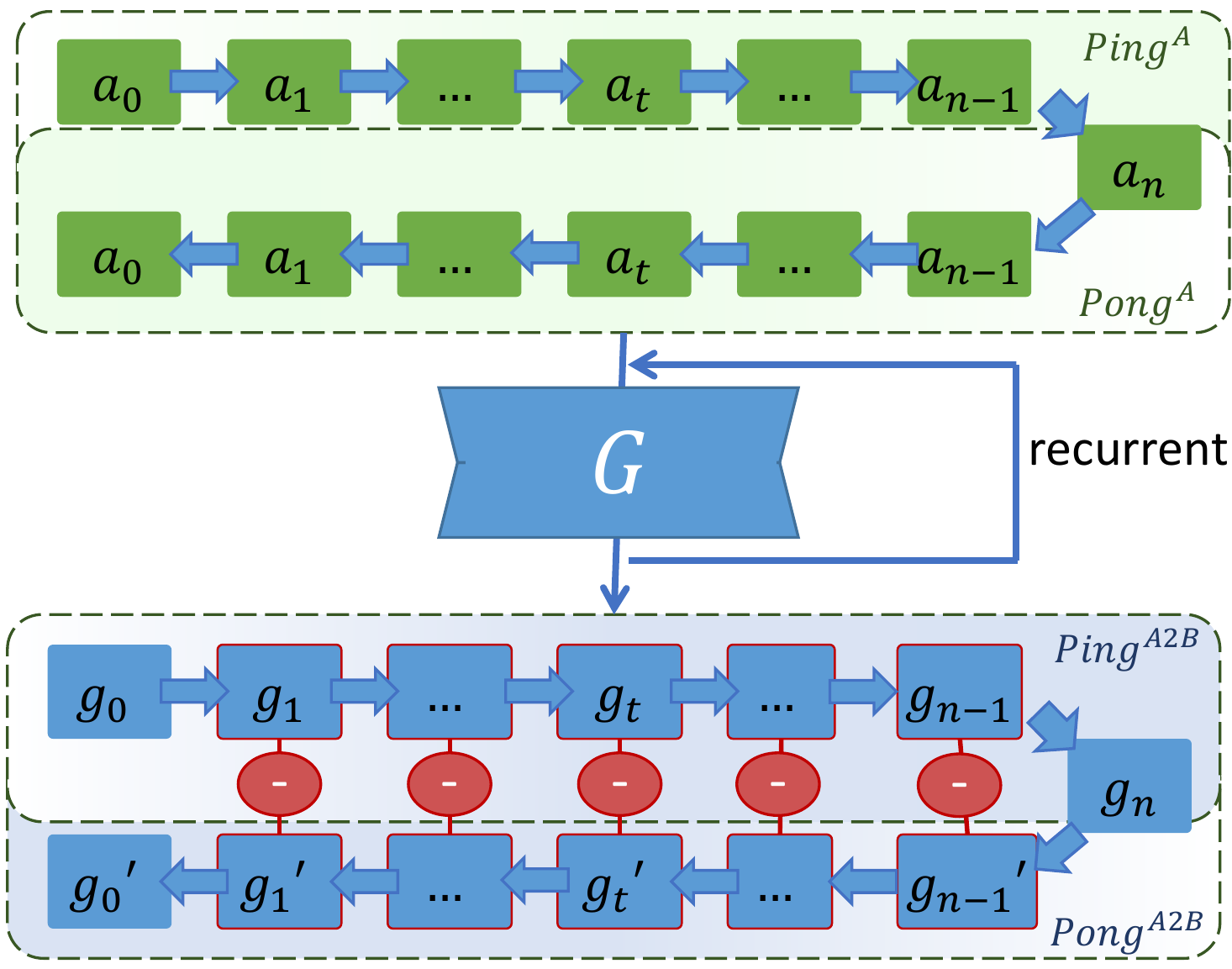}
\put(0,31){{ \color{black}{\textbf{c)}}}}
\end{overpic}
\end{minipage}
\caption{
{a) Result without \ppl{} loss.
The VSR network is trained with a recurrent frame-length of 10.
When inference on long sequences, frame 15 and latter frames of the foliage scene
show the drifting artifacts. \quad
b) Result trained with \ppl{} loss. These artifacts are removed successfully 
for the latter. \quad
c)
{When inferring a symmetric PP sequence with a forward pass (Ping) and its backward counterpart (Pong), our PP loss constrains the output sequence to be symmetric.
It reduces the $L^2$ distance between ${g}_{t}$ and ${g}_{t}'$, the corresponding frames in the forward and backward passes,
{shown via red circles with a minus sign.
The PP loss reduces drifting artifacts and improves temporal coherence.}}
}}
\label{fig:ping_pongcomparison}
\end{figure}

When relying on a previous output as input, i.e., for frame-recurrent architectures,
generated structures easily accumulate frame by frame.
In adversarial training, generators learn to heavily rely
on previously generated frames and can easily converge towards strongly reinforcing spatial features
over longer periods of time.
For videos, this especially occurs along directions of motion
and these solutions can be seen as a special form of temporal mode collapse,
{where the training converges to a mostly constant temporal signal as a sub-optimal, trivial equilibrium.}
We have noticed this issue in a variety of recurrent architectures, examples are shown in
\myreffig{fig:ping_pongcomparison} a) and the Dst version in \myreffig{fig:UVTCMP}.

While this issue could be alleviated by training with longer sequences,
{it is computationally expensive and can fail for even longer sequences,
as shown in \myrefapp{app:PPaug}.}
We generally want generators
to be able to work with sequences of arbitrary length for inference.
To address this inherent problem of recurrent generators,
we propose a new bi-directional ``Ping-Pong'' loss. 
For natural videos, a sequence with the forward order as well as its reversed counterpart
offer valid information. Thus, from any input of length $n$,
we can construct a symmetric {PP} sequence in form of
${a}_{1}, ...{a}_{n-1}, {a}_{n}, {a}_{n-1}, ...{a}_{1}$ as shown in \myreffig{fig:ping_pongcomparison} c).
When inferring this in a frame-recurrent manner,
the generated result should
not strengthen any invalid features from frame to frame.
Rather, the result should stay close to valid information 
and be symmetric, i.e.,
the forward result {\small ${g}_{t} = G({a}_{t}$, ${g}_{t-1})$} and
the one generated from the reversed part,
{\small ${g}_{t}' = G({a}_{t}$, ${g}_{t+1}')$},
should be identical.

Based on this observation, we train our networks with extended PP sequences
and	constrain the generated outputs from both ``legs'' to be the same
using the loss:
$\mathcal{L}_{pp} =\sum_{t=1}^{n-1} \left\| g_{t}-g_{t}{'} \right \|_{2} .$
Note that in contrast to the generator loss, the $L^2$ norm is a correct choice here:
We are not faced with multi-modal data where an $L^2$ norm would lead to undesirable
averaging, but rather aim to constrain the recurrent generator to its own, unique
version over time {without favoring smoothness}.
The \ppl{} terms provide constraints for short term consistency
via {\small $\left \| g_{n-1}-g_{n-1}{'} \right \|_{2}$}, while
terms such as {\small $\left \| g_{1}-g_{1}{'} \right \|_{2}$ }
prevent long-term drifts of the results.
{
This bi-directional loss formulation also helps to constrain ambiguities due to disocclusions that can occur in regular training scenarios.}

As shown in \myreffig{fig:ping_pongcomparison} b),
the \ppl{} loss successfully removes drifting artifacts
while appropriate high-frequency details are preserved.
In addition, it effectively
extends the training data set, and as such represents a
useful form of data augmentation.
{ A comparison is given in \myrefapp{app:PPaug}
to disentangle the effects of the augmentation
of PP sequences and the temporal constraints.
The results show that the temporal constraint is the key to reliably
suppressing the temporal accumulation of artifacts, achieving consistency,
and allowing models to infer much longer sequences than seen during training.
}

{The majority of related work for video generation focuses on network architectures.
Being orthogonal to architecture improvements, our work explores temporal self-supervision. The proposed spatio-temporal discriminator and the PP loss can be used in video generation tasks to replace simple temporal losses, e.g. the ones based on $L^2$ differences and warping.
In the following subsections, solutions for VSR and UVT are presented as examples in paired and unpaired data domains.}

\subsection{Network Architecture for VSR}
\label{sec:VSR}

\begin{figure}[t]
\centering
\begin{overpic}[width=0.24\linewidth]{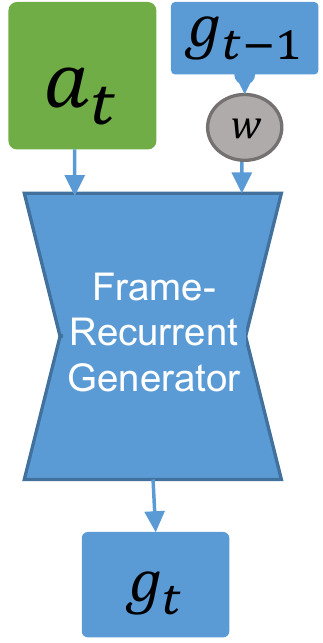}
\put(-5,1){{ \color{black}\textbf{{a)}}}}
\end{overpic}
\hspace{16pt}
\begin{overpic}[width=0.49\linewidth]{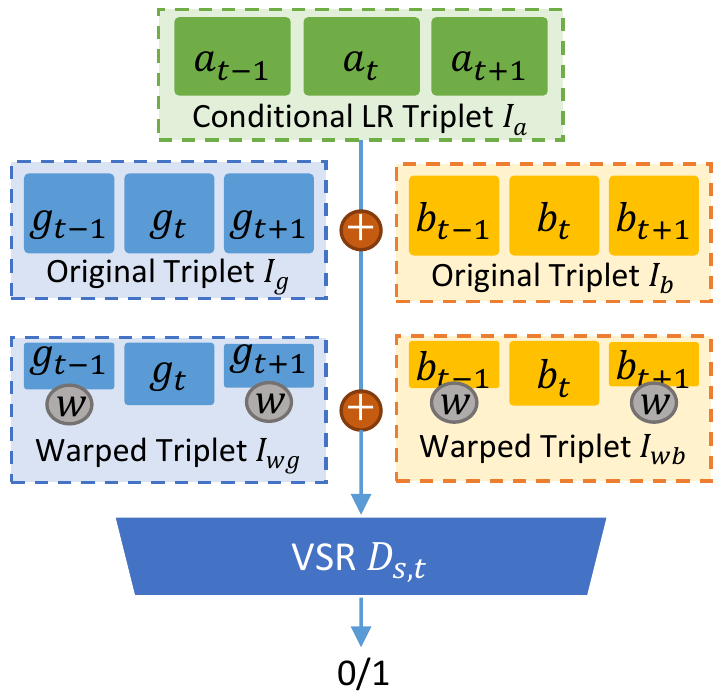}
\put(-5,1){{ \color{black}\textbf{{b)}}}}
\end{overpic}
\caption{a) The frame-recurrent VSR Generator. \quad b) {Conditional VSR $D_{s,t}$.}}
\label{fig:gnet}
\end{figure}

{For video super-resolution (VSR) tasks, the input domain contains LR frames while
the target domain contains high-resolution (HR) videos with more complex details and motions.
Since one pattern in low-resolution can correspond to multiple  structures in high-resolution,
VSR represents a multimodal problem that benefits from adversarial training.
In the proposed spatio-temproal adversarial training, we use a ResNet architecture for the VSR generator G.
Similar to previous work, an encoder-decoder structure is applied to $F$ for motion estimation.
We intentionally keep the generative part simple and in line with previous work, in order to demonstrate the advantages of the temporal self-supervision.}

The VSR discriminator $D_{s,t}$ should guide the generator to learn the
correlation between the conditional LR inputs and HR targets.
Therefore, three LR frames
$\text{I}_a =\{a_{t-1},a_t,a_{t+1}\}$
from the input domain are used as a conditional input.
The input of $D_{s,t}$ can be summarized as $ \text{I}_{s,t}^b = \{\text{I}_b, \text{I}_{wb}, \text{I}_a\}$ labelled as \textit{real} and the generated inputs
$ \text{I}_{s,t}^g = \{\text{I}_g, \text{I}_{wg}, \text{I}_a\}$ labelled as \textit{fake},
as shown in \myreffig{fig:gnet}.
We concatenate all triplets together.
In this way, the conditional $D_{s,t}$ will penalize $G$ if $\text{I}_g$ contains
less spatial details or unrealistic artifacts {in comparison} to $\text{I}_a,\text{I}_b$.
At the same time, temporal relationships between the generated images $\text{I}_{wg}$
and those of the ground truth $\text{I}_{wb}$ should match.
With our setup, the discriminator profits from the
warped frames to classify realistic and unnatural temporal changes, and
for situations where the motion estimation is less accurate,
the discriminator can fall back to the original, i.e. not warped, images.

\subsection{Network Architecture for UVT}
\label{sec:UVT}

\begin{figure}[t]
\centering
\begin{overpic}[width=0.47\linewidth]{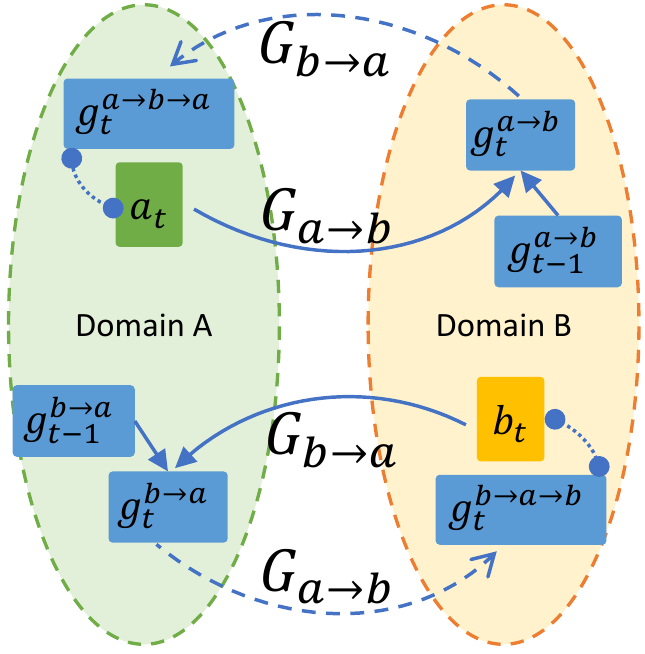}
\put(-2,0){{ \color{black}\textbf{{a)}}}}
\end{overpic}
\hfill
\begin{overpic}[width=0.49\linewidth]{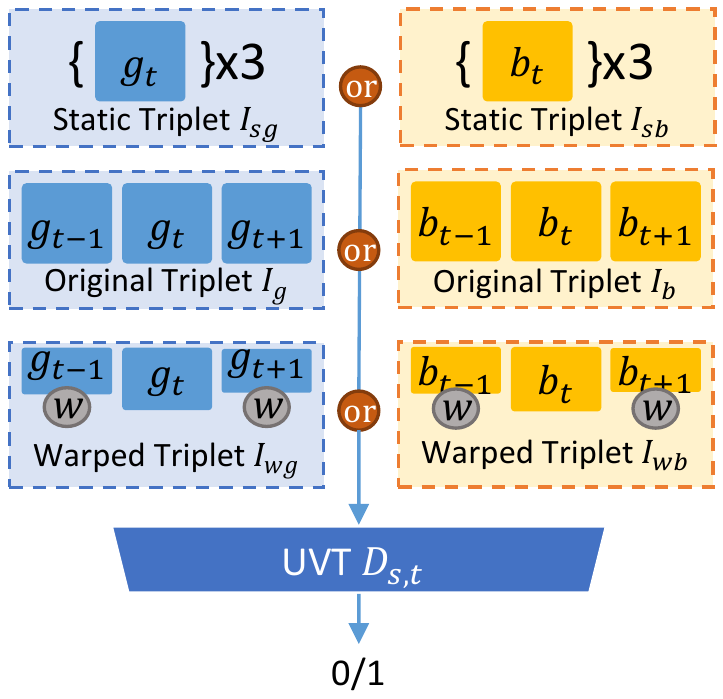}
\put(-2,0){{ \color{black}\textbf{{b)}}}}
\end{overpic}
\caption{a) The UVT cycle link formed by two recurrent generators.\quad b) {Unconditional UVT $D_{s,t}$}.}
\label{fig:UVTnet}
\end{figure}

While one generator is enough to map data from A to B for tasks such as VSR,
unpaired generation tasks require a second generator to establish cycle consistency
\citep{zhu2017cycleGAN}.
For the UVT task, we use two recurrent generators{, mapping from domain A to B and back.}
As shown in \myreffig{fig:UVTnet} a), given
$g^{a\rightarrow b}_t = \text{G}_\text{ab}( a_t,  W( g^{a\rightarrow b}_{t-1}, v_t ) )$,
we can use $a_t$ as the labeled data of
{\small $ g^{a\rightarrow b\rightarrow a}_t
= \text{G}_\text{ba}(g^{a\rightarrow b}_t, W( g^{a\rightarrow b\rightarrow a}_{t-1}, v_t ))$}
{to enforce consistency}.
An encoder-decoder structure is applied to UVT generators and $F$.

In UVT tasks, we demonstrate that the temporal cycle-consistency between different domains
can be established using the supervision of unconditional spatio-temporal discriminators.
This is in contrast to previous work which
focuses on the generative networks to form spatio-temporal cycle links.
Our approach actually yields improved results, as we will show below.
In practice, we found it crucial to ensure that generators first learn reasonable
spatial features, and only then improve
{their temporal correlation.}
Therefore, different to the $D_{s,t}$ of VST that always receives 3 concatenated triplets as an input,
the unconditional $D_{s,t}$ of UVT only takes one triplet at a time.
Focusing on the generated data,
the input for a single batch can either be a static triplet of
{\small $\text{I}_{sg}=\{g_t, g_t, g_t\}$},
the warped triplet $\text{I}_{wg}$, or the original triplet $\text{I}_g$.
The same holds for the reference data of the target domain,
as shown in \myreffig{fig:gnet} b).
Here, the warping is again performed via \fnet.
With sufficient but complex information contained in these triplets,
transition techniques are applied so that the network can
consider the spatio-temporal information step by step, i.e.,
we initially start with 100\mypercent{} static triplets $\text{I}_{sg}$ as the input.
{Then, over the course of training,
25\mypercent{} of them transit to $\text{I}_{wg}$ triplets with simpler temporal information,
with another 25\mypercent{} transition to  $\text{I}_{g}$ afterwards, leading to a (50\mypercent{},25\mypercent{},25\mypercent{}) distribution of triplets. Details of the transition calculations are given in \myrefapp{app:mcinD}.
} {Sample triplets are visualized in the supplemental web-page (Sec. 7).}

While non-adversarial training typically employs loss formulations with static goals,
the GAN training yields dynamic goals due to
discriminators  discovering
learning objectives over the course of the training run.
Therefore, their inputs have a strong influence on the training process and the final results.
Modifying the inputs in a controlled manner	can lead to different results
and substantial improvements if done correctly, as will be shown in \myrefsec{sec:evaluation}.

\subsection{Loss Functions}
\label{sec:loss_summary}

\paragraph{Perceptual Loss Terms}
As perceptual metrics, both {pre-trained} NNs~\cite{johnson2016perceptual,wang2018perceptual}
and {GAN} 
discriminators~\cite{you2018tempoGAN} were successfully
used in previous work.
Here, we use feature maps from a pre-trained VGG-19 network~\cite{simonyan2014very},
as well as $D_{s,t}$ itself.
In the VSR task, we can encourage the generator to produce features similar to the ground
truth ones by increasing the cosine similarity of their feature maps.
In UVT tasks without paired ground truth data, {the generators should match the distribution of features in the target domain. Similar to a style loss for traditional style transfer tasks \cite{johnson2016perceptual}, we thus compute}
the {$D_{s,t}$ feature correlations measured by the Gram matrix for UVT tasks.
The $D_{s,t}$  features contain both spatial and temporal
information and hence are especially well suited for the perceptual loss.}

{
\begin{figure*}[tp]
\begin{center}
\includegraphics[width=0.95\linewidth]{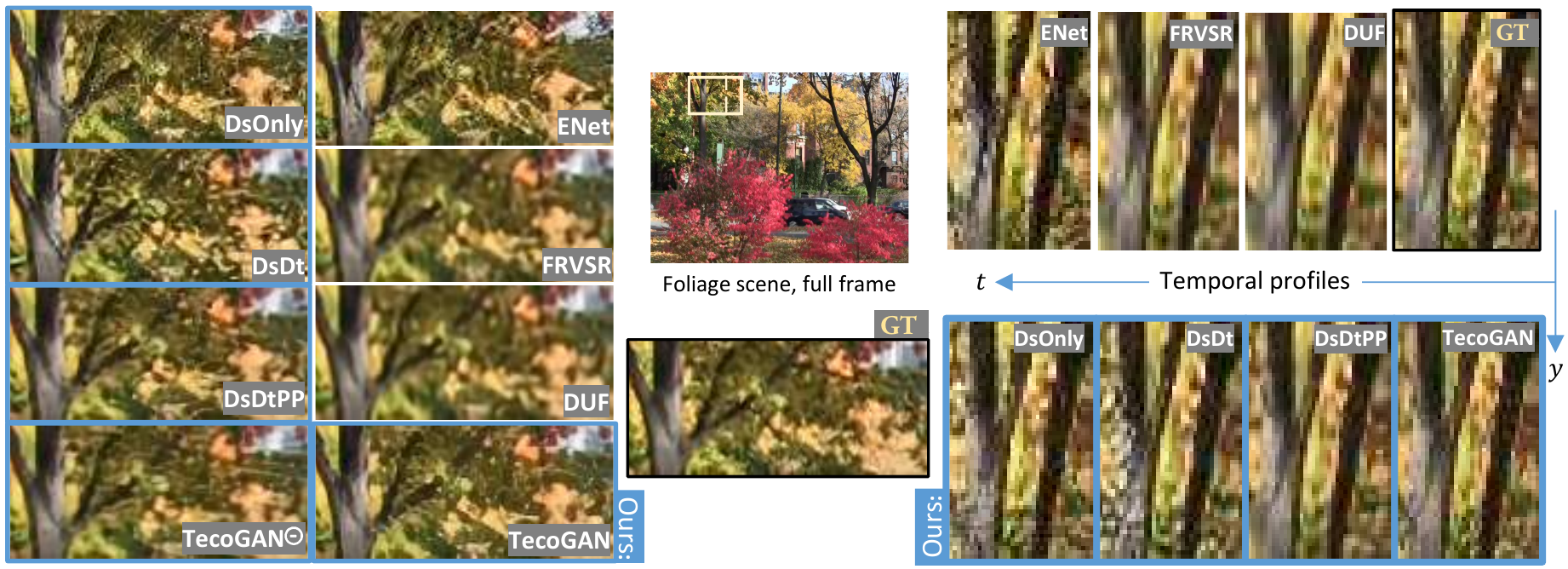}
\end{center} \vspace{-12pt}
\caption{\footnotesize{In VSR of the foliage scene, adversarial models
(ENet, DsOnly, DsDt, DsDtPP, \TecoGANsmall and TecoGAN) yield  better perceptual quality
than methods using $L^2$ loss (FRVSR and DUF).
In temporal profiles on the right,
DsDt, DsDtPP and TecoGAN show significantly less temporal discontinuities
compared to ENet and DsOnly.
The temporal information of our discriminators successfully suppresses these artifacts.
{Corresponding video clips can be found in Sec. 4.1-4.6 of \href{https://ge.in.tum.de/wp-content/uploads/2020/05/ClickMe.html}{the supplemental web-page} .}
}}
\label{fig:folCMP}
\end{figure*}
\begin{figure}[bp]
\centering
\includegraphics[width=0.75\linewidth]{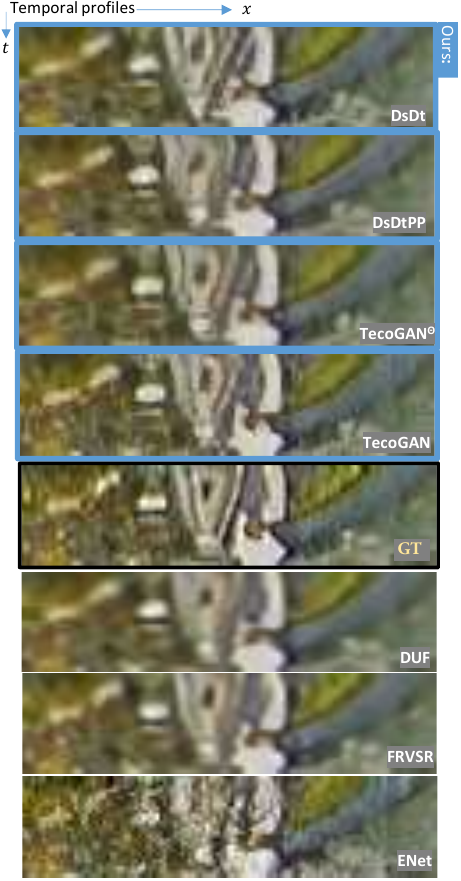} \vspace{-9pt}
\caption{\footnotesize{
VSR temporal profile comparisons of the calendar scene (time shown along y-axis),
{cf. Sec. 4.1-4.6 of the supplemental web-page.}
TecoGAN models lead to natural temporal progression, and
our final model closely matches the desired ground truth behavior
over time.}}
\label{fig:calCMP}
\end{figure}
}

\paragraph{Loss and Training Summary~}

We now explain how to integrate the spatio-temporal
discriminator into the paired and unpaired tasks.
We use a standard discriminator loss for
the $D_{s,t}$ of VSR and a least-square discriminator loss for the $D_{s,t}$ of UVT.
Correspondingly, a non-saturated $\mathcal{L}_{adv}$ is used for the $G$ and \fnet of VSR
and a least-squares one is used for the UVT generators. 
As summarized in \myreftab{tab:loss},
$G$ and \fnet are trained with
the mean squared loss $\mathcal{L}_\text{content}$, 
adversarial losses { $\mathcal{L}_{adv}$ }, perceptual losses
{ $\mathcal{L}_{\phi}$ }, the \ppl{} loss { $\mathcal{L}_{\text{PP}}$ },
and a warping loss $\mathcal{L}_{\text{warp}}$,
where again $g$, $b$ and $\Phi$ stand for generated samples, ground truth images and feature maps of VGG or $D_{s,t}$. We only show losses for the mapping from A to B for UVT tasks,
as the backward mapping simply mirrors the terms.
We refer to our full model for both tasks as {\em TecoGAN} below.
{The UVT data-sets are obtained from previous work~\cite{bansal2018recycle} and
each data domain has around 2400 to 3600 unpaired frames.
For VSR, we download 250 short videos with 120 frames each from \url{Vimeo.com}.
In line with other VSR projects, we down-sample these frames by a factor of 2 to
get the ground-truth HR frames.
Corresponding LR frames are achieved by applying a Gaussian blur and sampling every fourth pixel.
A Gaussian blur step is important to mimic the information loss due to the
camera sensibility in a real-life capturing scenario.
Although the information loss is complex and not unified, a {Gaussian kernel with} a standard deviation of 1.5 is commonly used for a super-resolution factor of 4.
}
Training parameters and details are given in \myrefapp{app:training}.

{
\begin{table}[h]\small
\begin{center}
\caption{{Summary of loss terms.}}\label{tab:loss}
\vspace{-14pt}
\setlength{\tabcolsep}{3pt}
\setlength\extrarowheight{0.1pt}
\begin{tabular}{c|c} 
\hline
\multicolumn{2}{l}{$ \mathcal{L}_{D_{s,t}} $ for }\\\hline
{VSR, $D_{s,t}$} &
{$
- \mathbb{E}_{b\sim p_{\text{b}}(b)}[\log D(\text{I}_{s,t}^b)] -\mathbb{E}_{a\sim p_{\text{a}}(a)}[\log (1-D(\text{I}_{s,t}^g))]
$}
\\\hline
{UVT, $D_{s,t}^{b}$} &
{$
\mathrm{E}_{b \sim p(b)} [D(\text{I}_{s,t}^b)-1]^2 + \mathrm{E}_{a \sim p(a)} [D(\text{I}_{s,t}^g)]^2
$}
\\\hline
\end{tabular}
\newline
\setlength{\tabcolsep}{0.1pt}
\setlength\extrarowheight{0.1pt}
\begin{tabular}{c|c|c} 
\hline
{Loss for} & {VSR, G \& \fnet} & {UVT, $G_{ab}$} \\[2pt]\hline
{$ \mathcal{L}_{G,F}$} & \multicolumn{2}{c}{ $
\lambda_{w} \mathcal{L}_\text{warp} + \lambda_{p} \mathcal{L}_\text{PP}
+ \lambda_{a} \mathcal{L}_\text{adv} + \lambda_{\phi} \mathcal{L}_\phi
+ \lambda_{c} \mathcal{L}_\text{content}
$}\\[2pt]\hline
{$\mathcal{L}_\text{warp}$} &
\multicolumn{2}{c}{
$\myavg \left \| a_t - W( a_{t-1}, \text{F}(a_{t-1}, a_t))\right \|_{2} $
} \\[2pt]\hline
{$\mathcal{L}_\text{PP}$} &
\multicolumn{2}{c}{
$\sum_{t=1}^{n-1} \left\| g_{t}-g_{t}{'} \right \|_{2}$
}\\[2pt]\hline
{$\mathcal{L}_\text{adv}$} &
{$ - \mathbb{E}_{a\sim p_{\text{a}}(a)} [\log D_{s,t}(\text{I}_{s,t}^g)]$  }
&
{$- \mathbb{E}_{a\sim p_{\text{a}}(a)} [D^b_{s,t}(\text{I}_{s,t}^{g^{a\rightarrow b}})]^2$ }
\\[2pt]\hline
{$\mathcal{L}_\phi$} &   1.0 - $\frac {\Phi(\text{I}_{s,t}^g) * \Phi(\text{I}_{s,t}^b) }{\left \| \Phi(\text{I}_{s,t}^g)\right \| * \left \|\Phi(\text{I}_{s,t}^b)\right \|}$ &
{$\left \| GM(\Phi(\text{I}_{s,t}^g)) - GM(\Phi (\text{I}_{s,t}^b))\right \|_{2}$}
\\[3pt]\hline
{$\mathcal{L}_\text{content}$} &
$\left \| g_t-b_t \right \|_{2}$ &
{\footnotesize
$\left \| g^{a\veryshortarrow b \veryshortarrow a}_t-a_t \right \|_{2} +
\left \| g^{b\veryshortarrow a \veryshortarrow b}_t-b_t \right \|_{2} $
}
\\[2pt]\hline
\end{tabular}
\end{center}
\end{table}
}

\section{Analysis and Evaluation of Learning Objectives}\label{sec:evaluation}

In the following section,
we illustrate the effects of temporal supervision using two ablation studies.
In the first one, models trained with ablated loss functions
show how $\mathcal{L}_\text{adv} $ and $\mathcal{L}_\text{PP} $ change the overall learning objectives.
Next, full UVT models are trained with different $D_{s,t}$ inputs.
This highlights how differently the corresponding
discriminators converge to different spatio-temporal equilibriums and the general
importance of providing suitable data distributions from the target domain.
While we provide qualitative and quantitative evaluations below, we also refer the reader to our supplemental
material which contains a \href{https://ge.in.tum.de/wp-content/uploads/2020/05/ClickMe.html}{web-page} with
video clips that more clearly highlight the temporal differences.

\subsection{Loss Ablation Study}

Below we compare variants of our full TecoGAN model to
EnhanceNet (ENet)~\cite{sajjadi2017enhancenet}, FRVSR~\cite{sajjadi2018FRVSR}, and
DUF~\cite{jo2018deep} for VSR.
CycleGAN~\cite{zhu2017cycleGAN} and RecycleGAN~\cite{bansal2018recycle} are compared for UVT.
Specifically, ENet and CycleGAN represent
state-of-the-art single-image adversarial models without temporal information,
FRVSR and DUF are state-of-the-art VSR methods without adversarial losses,
and RecycleGAN is a spatial adversarial model with a
prediction network learning the temporal evolution.

\begin{figure*}[t]\footnotesize
\begin{center}
\begin{overpic}[width=0.16\linewidth]{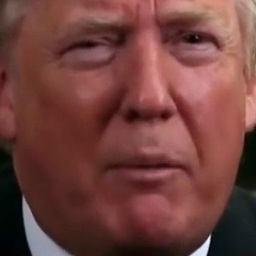}
\put(-2,4){{ \color{white}\textbf{{Input}}}}
\end{overpic}
\begin{overpic}[width=0.16\linewidth]{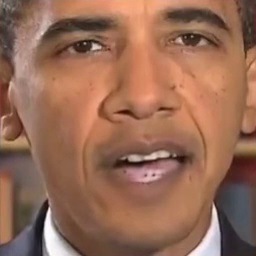}
\put(-2,4){{ \color{white}{\textbf{CycleGAN}}}}
\end{overpic}
\begin{overpic}[width=0.16\linewidth]{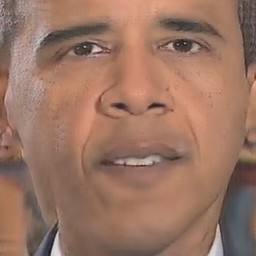}
\put(-2,4){{ \color{white}{\textbf{DsOnly}}}}
\end{overpic}
\begin{overpic}[width=0.16\linewidth]{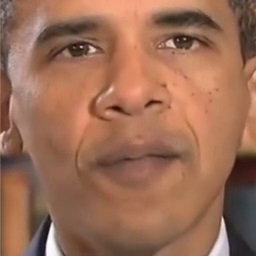}
\put(-2,4){{ \color{white}{\textbf{RecycleGAN}}}}\end{overpic}
\begin{overpic}[width=0.16\linewidth]{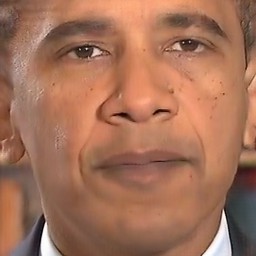}
\put(-2,4){{ \color{white}\textbf{
STC-V2V
}}}
\end{overpic}
\begin{overpic}[width=0.16\linewidth]{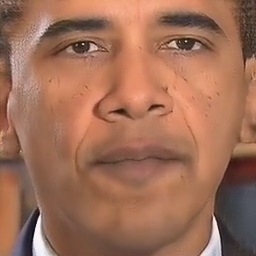}
\put(-2,4){{ \color{white}{\textbf{TecoGAN}}}}\end{overpic}\\
\begin{overpic}[width=0.16\linewidth]{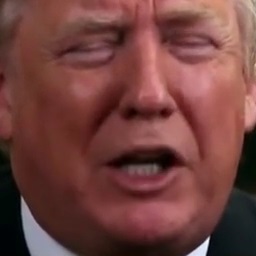}
\put(-2,4){{ \color{white}{\textbf{Input}}}}\end{overpic}
\begin{overpic}[width=0.16\linewidth]{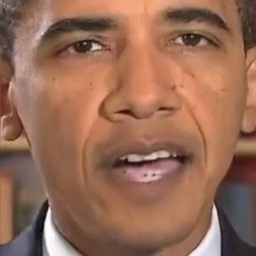}
\put(-2,4){{ \color{white}{\textbf{CycleGAN}}}}\end{overpic}
\begin{overpic}[width=0.16\linewidth]{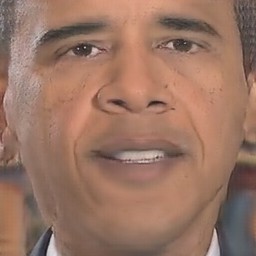}
\put(-2,4){{ \color{white}{\textbf{DsOnly}}}}\end{overpic}
\begin{overpic}[width=0.16\linewidth]{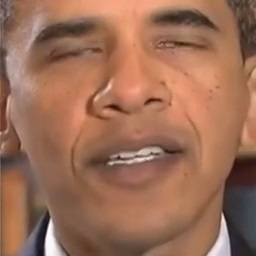}
\put(-2,4){{ \color{white}{\textbf{RecycleGAN}}}}\end{overpic}
\begin{overpic}[width=0.16\linewidth]{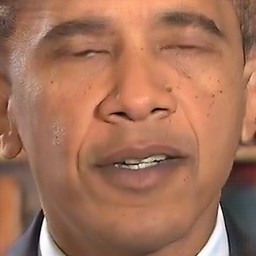}
\put(-2,4){{ \color{white}\textbf{{
STC-V2V
}}}}
\end{overpic}
\begin{overpic}[width=0.16\linewidth]{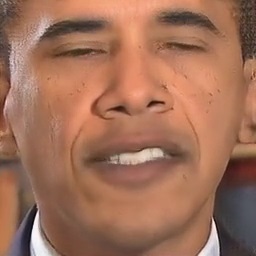}
\put(-2,4){{ \color{white}{\textbf{TecoGAN}}}}\end{overpic}\\
\begin{overpic}[width=0.16\linewidth]{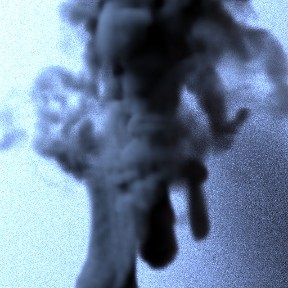}
\put(0,4){{ \color{black}{\textbf{Input}}}}\end{overpic}
\begin{overpic}[width=0.16\linewidth]{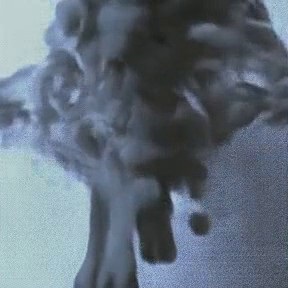}
\put(0,12){{ \color{black}{\textbf{Cycle-}}}}
\put(0,4){{ \color{black}{\textbf{GAN}}}}\end{overpic}
\begin{overpic}[width=0.16\linewidth]{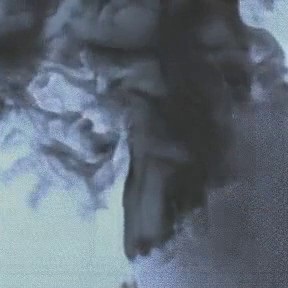}
\put(0,4){{ \color{black}{\textbf{DsOnly}}}}\end{overpic}
\begin{overpic}[width=0.16\linewidth]{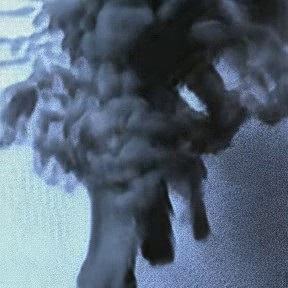}
\put(0,4){{ \color{black}{\textbf{Dst}}}}\end{overpic}
\begin{overpic}[width=0.16\linewidth]{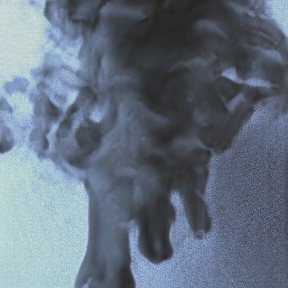}
\put(0,4){{ \color{black}{\textbf{DsDtPP}}}}\end{overpic}
\begin{overpic}[width=0.16\linewidth]{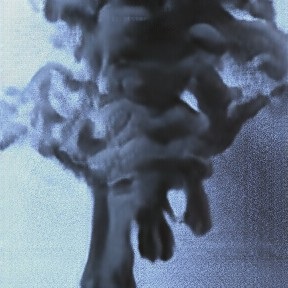}
\put(0,12){{ \color{black}{\textbf{Teco-}}}}
\put(0,4){{ \color{black}{\textbf{GAN}}}}\end{overpic}\\
\end{center}
\vspace{-9pt}
\caption{{\footnotesize When learning a mapping between Trump and Obama,
the CycleGAN model gives good spatial features but collapses to essentially static outputs of Obama.
It manages to transfer facial expressions
back to Trump using tiny differences encoded in its Obama outputs,
without understanding the cycle-consistency between the two domains.
Being able to establish the correct temporal cycle-consistency between domains, ours, RecycleGAN and {STC-V2V} 
can generate correct blinking motions, shown in Sec. 4.7 of the supplemental web-page.
Our model outperforms the latter two in terms of coherent detail that is generated.
Obama and Trump video courtesy of the White House (public domain).
}}\label{fig:UVTCMP} 
\end{figure*}

\paragraph{{Video Super-Resolution}}
For VSR, we first train a {\em DsOnly} model that uses a frame-recurrent $G$ and \fnet with a
VGG loss and only the regular spatial discriminator.
Compared to ENet, which exhibits strong incoherence due to the lack of temporal information,
DsOnly improves temporal coherence thanks to the frame-recurrent connection,
but there are noticeable high-frequency changes between frames.
The temporal profiles of DsOnly in \myreffig{fig:folCMP} and \ref{fig:calCMP},
correspondingly contain sharp and broken lines.

When adding a temporal discriminator in addition to the spatial one ({\em DsDt}),
this version generates more coherent results,
and its temporal profiles are sharp and coherent.
However, DsDt often produces the drifting artifacts discussed in \myrefsec{sec:method},
as the generator learns to reinforce existing
details from previous frames to fool $D_s$ with sharpness,
and satisfying $D_t$ with good temporal coherence in the form of persistent detail.
While this strategy works for generating short sequences during training,
the strengthening effect 
can lead to very undesirable artifacts for long-sequence inferences.

By adding the self-supervision for long-term temporal consistency $\mathcal{L}_{pp}$,
we arrive at the {\em DsDtPP} model,
which effectively suppresses these drifting artifacts with an improved temporal coherence.
In \myreffig{fig:folCMP} and \myreffig{fig:calCMP},
DsDtPP results in continuous yet detailed temporal profiles
without streaks from temporal drifting. Although DsDtPP generates good results,
it is difficult in practice to balance the generator and the two discriminators.
The results shown here were achieved only after numerous runs manually tuning the weights of the different loss terms.
By using the proposed $D_{s,t}$ discriminator instead, we get a first complete model
for our method, denoted as {\em \TecoGANsmall}.
This network is trained with a
discriminator that achieves an excellent quality with an effectively
halved network size, as illustrated on the right of \myreffig{fig:metrics}.
The single discriminator correspondingly leads to a significant reduction in resource usage.
Using two discriminators requires ca. 70\mypercent{} more GPU memory, and leads to a reduced training performance by ca. 20\mypercent{}.
The \TecoGANsmall model yields similar perceptual and temporal quality
to DsDtPP with a significantly faster and more stable training.

Since the \TecoGANsmall model requires less training resources,
we also trained a larger generator with 50\mypercent{} more weights.
In the following, we will focus on this larger single-discriminator architecture with \ppl{} loss as our full {\em TecoGAN} model for VSR.
Compared to the \TecoGANsmall model, it can generate more details,
and the training process is more stable, indicating that the larger generator and
$D_{s,t}$ are more evenly balanced.
Result images and temporal profiles are shown in \myreffig{fig:folCMP} and \myreffig{fig:calCMP}.
Video results are shown in Sec. 4 of the supplemental web-page. 

\paragraph{{Unpaired Video Translation}}
We carry out a similar ablation study for the UVT task.
Again, we start from a single-image GAN-based model, a {\em CycleGAN} variant
which already has two pairs of spatial generators and discriminators.
Then, we train the {\em DsOnly} variant by adding {flow estimation via} \fnet and extending the spatial generators
to frame-recurrent ones.
{By augmenting the two discriminators to use the triplet inputs
proposed in \myrefsec{sec:method}, we arrive at the {\em Dst} model with spatio-temporal discriminators,}
which does not yet use the \ppl{} loss.
{By adding the \ppl{} loss we complete the TecoGAN model for UVT.}
{Although UVT tasks substantially differ from VSR tasks,
the comparisons in {\myreffig{fig:UVTCMP}} and Sec. 4.7 of our supplemental web-page illustrate that UVT tasks profit from the proposed approach in a very similar manner to VSR}.

{We use renderings} of 3D fluid simulations of rising smoke as our unpaired training data.
These simulations are generated with randomized numerical simulations using a resolution of $64^3$ for domain A and $256^3$ for domain B, and both are visualized with images of size $256^2$.
Therefore, video translation from domain A to B is a tough task, as the latter contains significantly more turbulent and small-scale motions.
With no temporal information available,
the CycleGAN variant generates HR smoke that strongly flickers.
The DsOnly model offers better temporal coherence by relying on its frame-recurrent input,
but it learns a solution that largely ignores the current input
and fails to keep reasonable
spatio-temporal cycle-consistency links between the two domains.
On the contrary, our $D_{s,t}$ enables the Dst model to learn the correlation between
the spatial and temporal aspects, thus improving the cycle-consistency.
However, without $\mathcal{L}_{pp}$, the Dst model (like the DsDt model of VSR) reinforces detail over time in an undesirable way. This manifests itself as inappropriate smoke density in empty regions.
Using our full TecoGAN model which includes $\mathcal{L}_{pp}$, yields the best results, with detailed smoke structures and very good spatio-temporal cycle-consistency.

For comparison,
a DsDtPP model with a larger number of networks,
i.e. four discriminators,
two frame-recurrent generators and the \fnet,
is trained. By weighting the temporal adversarial losses from Dt with 0.3
and the spatial ones from Ds with 0.5, we arrived at a balanced training run.
Although this model performs similarly to the TecoGAN
model on the smoke dataset,
the proposed spatio-temporal  $D_{s,t}$ architecture represents
a more preferable choice in practice, as it learns a natural
balance of temporal and spatial components by itself, and requires fewer resources.
Continuing along this direction, it will be interesting future work to evaluate variants,
such as a shared $D_{s,t}$ for both domains, i.e. a multi-class classifier network.
Besides the smoke dataset, an ablation study for the Obama and Trump dataset from
\myreffig{fig:UVTCMP} shows a very similar behavior, as can be seen in the supplemental web-page (Sec. 4.7).

\begin{figure}[b]
\includegraphics[width=0.98\linewidth]{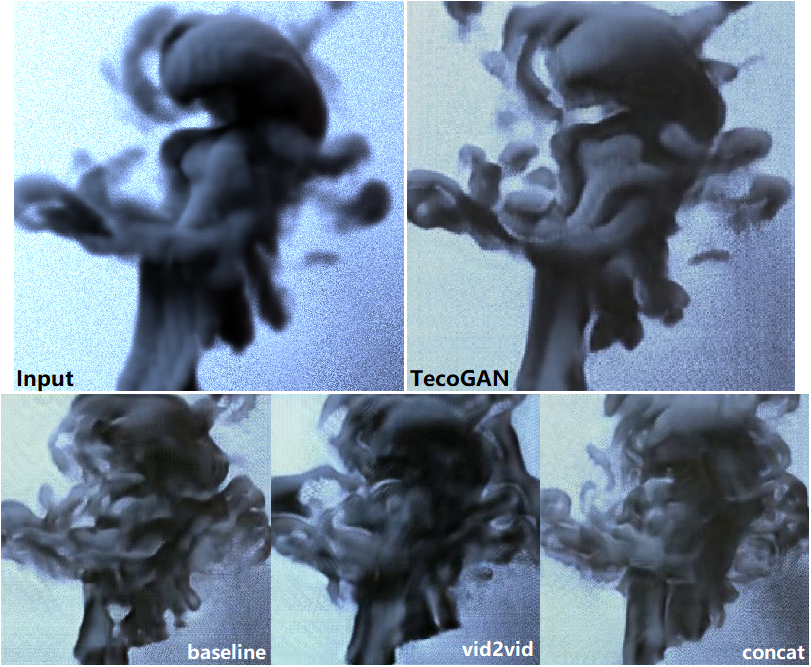}\vspace{-6pt}
\caption{{
Adversarial training arrives at different equilibriums when discriminators use different inputs.
The baseline model (supervised on original triplets) and the vid2vid variant (supervised on original triplets and estimated motions) fail to learn the complex temporal dynamics of a high-resolution smoke.
The warped triplets improve the result of the concat model and the full TecoGAN model performs better spatio-temporally.
Video comparisons are shown in Sec 5. of the supplemental web-page.
}}
\label{fig:abla2}
\end{figure}
\begin{figure}[b]
\begin{overpic}[width=\linewidth]{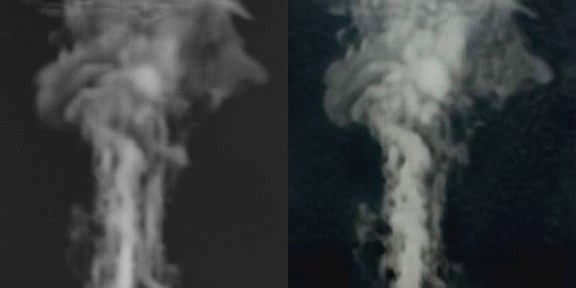}
\put(0,46){{ \color{white}{\textbf{Input}}}}
\put(79,46){{ \color{white}{\textbf{TecoGAN}}}}\end{overpic}
\vspace{-9pt}
\caption{Video translations between renderings of smoke simulations and real-world captures for smokes.}
\label{fig:UVTmore}
\end{figure}

{
\subsection{Spatio-temporal Adversarial Equilibriums}
Our evaluation so far highlights that temporal adversarial learning is crucial for achieving spatial detail that is coherent over time for VSR, and for {enabling the generators to} learn the spatio-temporal correlation between domains in UVT. Next, we will shed light on the complex
{spatio-temporal adversarial learning objectives}
by varying the information provided to the discriminator network. In the following tests,
{shown in \myreffig{fig:abla2} and Sec. 5 of the supplemental document,}
$D_{s,t}$ networks are identical apart from changing inputs, and we focus on the smoke dataset.}

In order to learn the spatial and temporal features of the target domain as well
as their correlation, the simplest input for $D_{s,t}$
consists of only the original, unwarped
triplets, i.e. $\{\text{I}_g$ or $\text{I}_b\}$.
Using these, we train a {\em{baseline}} model, which yields a sub-optimal quality: it lacks sharp spatial structures and contains coherent but dull motions.
Despite containing the full information, these input triplets prevent $D_{s,t}$ from
{providing} the desired supervision.
For paired video translation tasks, the {\em{vid2vid}} network achieves improved temporal coherence
by using a video discriminator to supervise the output sequence  conditioned with the ground-truth motion.
With no ground-truth data available, we train a vid2vid variant by
using the estimated motions and original triplets, i.e
$\{\text{I}_g + F(g_{t-1},g_t) + F(g_{t+1},g_t) $ or $ \text{I}_b+ F(b_{t-1},b_t) + F(b_{t+1},b_t)\}$,
as the input for $D_{s,t}$.
However, the result do not significantly improve. The motions are only partially reliable, and hence don't help for the difficult unpaired translation task. 
{Therefore}, the discriminator still fails to fully correlate spatial and temporal features.

\begin{figure}[tp]
\centering 
\includegraphics[width=0.9\linewidth]{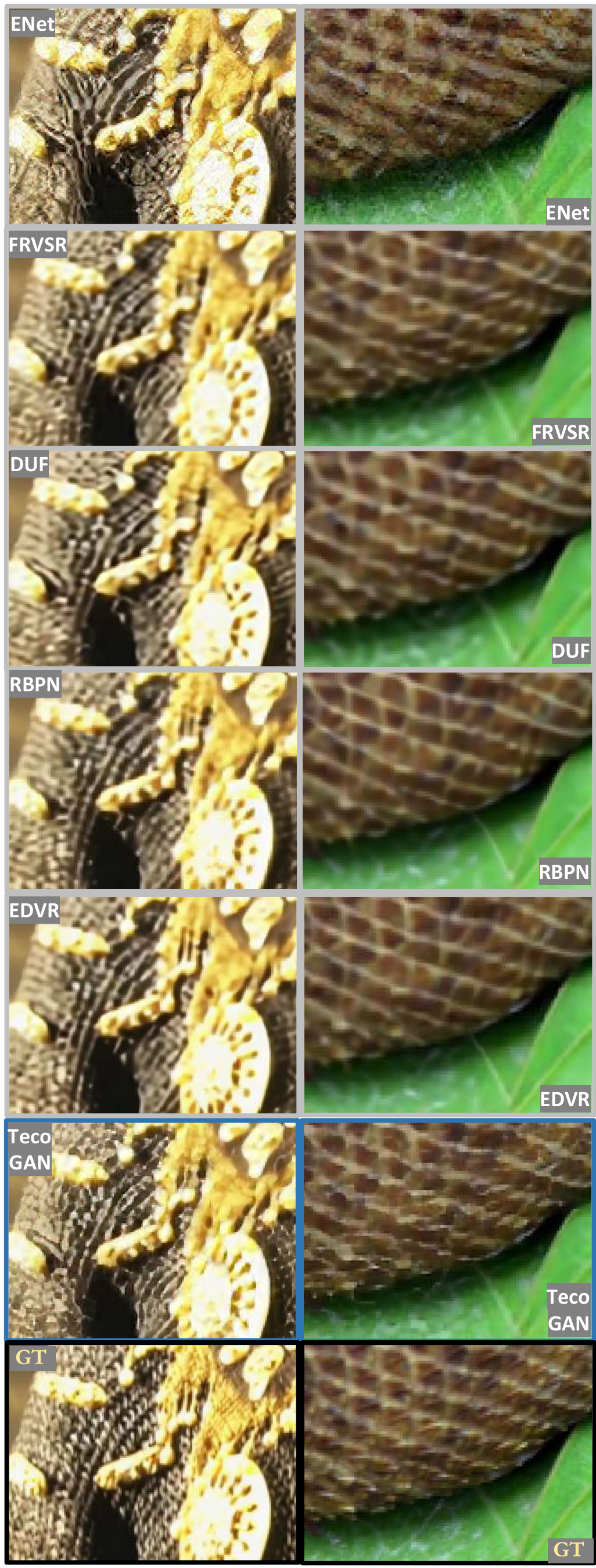}
\caption{Additional VSR comparisons, with videos in Sec 2 of the supplemental web-page. The TecoGAN model generates sharp details in both scenes.}
\label{fig:vid2}
\end{figure}

\begin{figure*}[p]
\centering 
\includegraphics[width=0.88\linewidth]{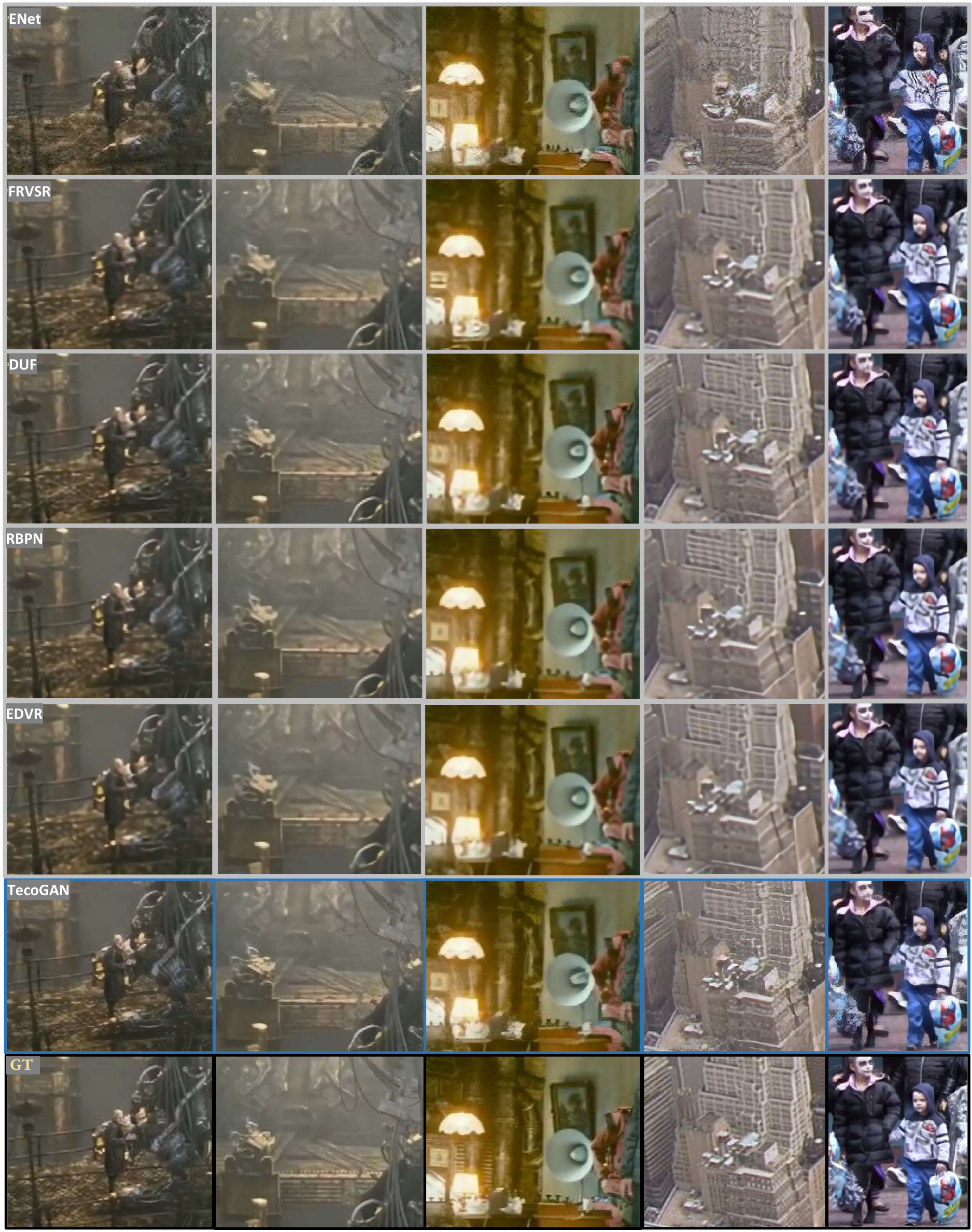}
\caption{{ \footnotesize Detail views of the VSR results of ToS scenes (first three columns) and Vid4 scenes (two right-most columns)
{generated with different methods: from top to bottom. ENet~\citep{sajjadi2017enhancenet}, FRVSR~\citep{sajjadi2018FRVSR}, DUF~\citep{jo2018deep}, {RBPN}~\citep{haris2019recurrent}, {EDVR}~\citep{wang2019edvr}, TecoGAN, and the ground truth.
Tears of Steel~(ToS)~ movie {(CC) Blender Foundation $|$ mango.blender.org}. }}
}\label{fig:suppresults1}
\end{figure*}

We then train a third model, {\em{concat}}, using the original triplets and the warped ones, i.e. $\{\text{I}_g+\text{I}_{wg}$ or $\text{I}_b+\text{I}_{wb}\}$. 
In this case, the model learns to generate more spatial details with a more vivid motion.
I.e., the improved temporal information from the warped triplets gives the discriminator important cues.
However, the motion still does not fully resemble the target domain.
We arrive at our final {\em{TecoGAN}} model for UVT by controlling the
composition of the input data: as outlined above, we first
provide only static triplets $\{\text{I}_{sg}$ or $\text{I}_{sb}\}$,
and then apply the transitions of
warped triplets $\{\text{I}_{wg}$ or $\text{I}_{wb}\}$, and original triplets $\{\text{I}_g$ or $\text{I}_b\}$ over the course of training.
In this way, the network can first learn to extract spatial features and build on them to establish temporal features.
{Finally, discriminators learn features about the correlation of spatial and temporal content by analyzing the original triplets and provide gradients such that the generators
{learn to use the motion information from the input and } 
establish a correlation between the motions in the two unpaired domains.}
{Consequently,} the discriminator, despite receiving only a single triplet at once, can guide the generator to produce detailed structures that move coherently.

\begin{figure*}[t]
\centering
\begin{overpic}[width=0.59 \linewidth]{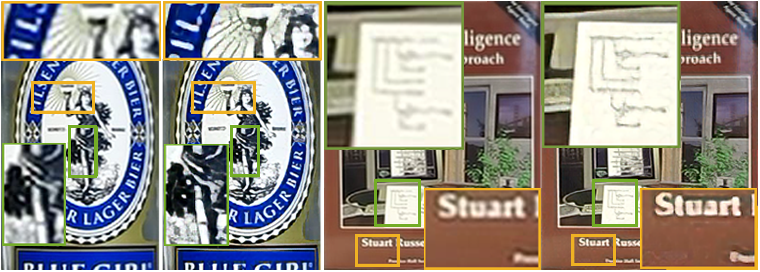}
\put(2,-2){\footnotesize{ \color{black}{\cite{liao2015video}}}}
\put(28,-2){\footnotesize{ \color{black}{Ours}}}
\put(49,-2){\footnotesize{ \color{black}{\cite{liao2015video}}}}
\put(80,-2){\footnotesize{ \color{black}{Ours}}}
\end{overpic}
\begin{overpic}[width=0.38 \linewidth]{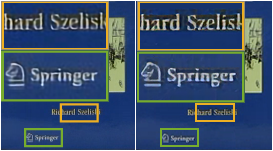}
\put(12,-3){\footnotesize{ \color{black}\cite{tao2017spmc}}}
\put(70,-3){\footnotesize{ \color{black}{Ours}}}
\end{overpic}
\caption{VSR comparisons for different
captured images {in order to compare to previous work  \cite{liao2015video,tao2017spmc}.}
}
\label{fig:real}
\end{figure*}

{\setlength{\tabcolsep}{10pt}
\begin{figure*}[!t]
\begin{minipage}{0.98\linewidth}
\centering
\begin{overpic}[width=\textwidth]{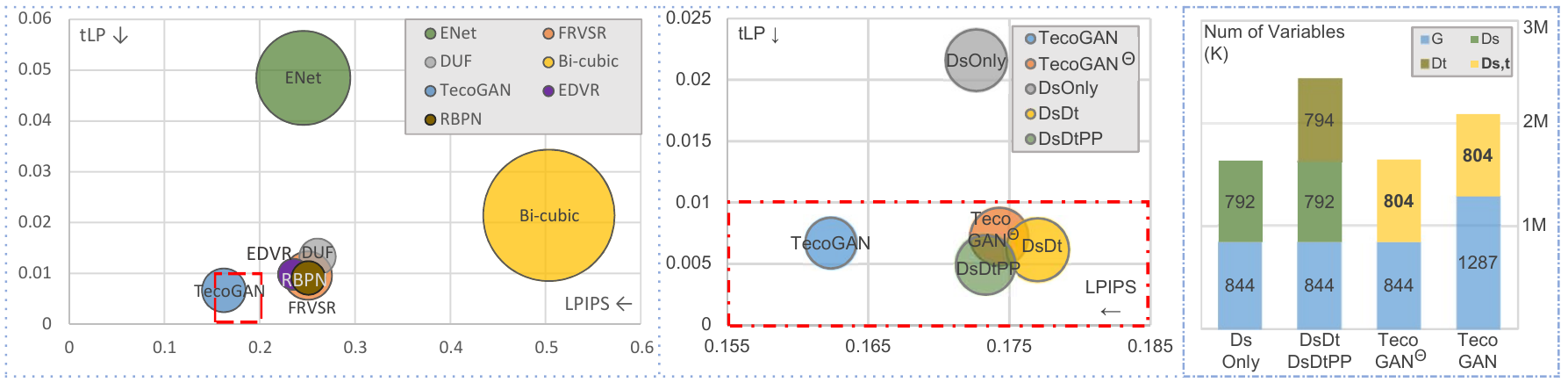}
{\put(1,1){a)}
\put(43,1){b)}
\put(75.6,1){c)}
}
\end{overpic}
\vspace{-16pt}
\caption{{Visual summary of VSR models. \quad
a) LPIPS (x-axis) measures
spatial detail and temporal coherence is measured by \selflp{} (y-axis)
and \ofmae{} (bubble size with smaller as better). \quad
b) The red-dashed-box region of a),
containing our ablated models. \quad
c) The network sizes.
}}
\label{fig:metrics}
\end{minipage}
\begin{minipage}{0.98\linewidth}
\begin{center}
\vspace{6pt}
\captionof{table}{{Averaged VSR metric evaluations for the \textit{Vid4} data set
with the following metrics,
PSNR: pixel-wise accuracy. 
LPIPS~{(AlexNet)}: perceptual distance to the ground truth. 
T-diff: pixel-wise differences of warped frames.
\ofmae{}: pixel-wise distance of estimated motions.
\selflp{}: perceptual distance between consecutive frames.
{User study: Bradley-Terry scores~\cite{bradley1952rank}.
{Performance is averaged over 500 images up-scaled from 320x134 to 1280x536.}
More details can be found in \myrefapp{app:metrics}
and Sec. 3 of the supplemental web-page.}
}}\label{tab:metrics} \vspace{-6pt}
\begin{tabular}{l|c|c||c|c|c||c||c|c} 
\hline
\rule{0pt}{11pt}
{Methods} & {PSNR$\uparrow$} & 
{$^{\text{LPIPS}\downarrow} _{\scriptstyle \times 10} $} &
{$^{\text{T-diff}\downarrow} _{\scriptstyle \times 100} $} &
{$^{\text{\ofmae{}}\downarrow} _{\scriptstyle \times 10} $} & {$^{\text{\selflp{}}\downarrow} _{\scriptstyle \times 100} $} & {$^{\text{User}} _{\text{Study}}\uparrow$} &
{$^{\text{Model}} _{\text{Size(M)}}\downarrow$} &
{$^{\text{Processing}} _{\text{Time(ms/frame)}}\downarrow$} \\
\hline
\rowcolor{lightcyan}
{DsOnly} & 24.14 & 
1.727 & 6.852 & 2.157 & 2.160 & - & -& -\\\hline 
\rowcolor{lightcyan}
{DsDt} & 24.75 & 
1.770 & 5.071 & 2.198 & 0.614 & - & -& -\\\hline 
\rowcolor{lightcyan}
{DsDtPP} & 25.77 & 
1.733 & 4.369 & 2.103 & { 0.489 } & -& -& -\\\hline 
\rowcolor{lightcyan}
{{\TecoGANsmall}} & 25.89 &
1.743 & 4.076 & 2.082 & 0.718 & -&
{{0.8(G)+1.7(F)}} & 37.07 \\\hline 
\rowcolor{lightcyan}
\textbf{{TecoGAN}} & 25.57 & 
\textbf{1.623} & 4.961 & 1.897 & \textbf{0.668} & \textbf{3.258}&
{{1.3(G)+1.7(F)}} & 41.92\\ 
\hline
\hline
{ENet} & 22.31 
& 2.458 & 9.281 & 4.009 & 4.848 & 1.616 & -& -\\
\hline
{FRVSR} & 26.91 
& 2.506 & 3.648 & 2.090 & 0.957 & 2.600 &
{{0.8(SRNet)+1.7(F)}} & 36.95\\ 
\hline
{DUF} & \textbf{27.38} 
& 2.607 & 3.298 & {1.588} & 1.329 & 2.933 &
{{6.2}} & 942.21\\ 
\hline
{Bi-cubic} & 23.66 
& 5.036 & 3.152 & 5.578 & 2.144 & 0.0 & -& -\\ 
\hline
{RBPN} & 27.15 
& 2.511 & - & 1.473 & 0.911 & - & 					12.7 & 510.90\\ 
\hline
{EDVR
} & 27.34 
& 2.356 & - & \textbf{1.367} & 0.982 & - &
20.7 & 299.71\\ 
\hline 
\end{tabular}
\end{center}
\end{minipage}
\end{figure*}

\section{Results and Metric Evaluation}
\label{sec:results}

For the VSR task, we test our model on a wide range of
video data, including the widely used Vid4 dataset
shown in \myreffig{fig:folCMP}, \ref{fig:calCMP} and \ref{fig:suppresults1},
detailed scenes from the movie Tears of Steel~(ToS)~\shortcite{tearsofsteel}
shown in \myreffig{fig:suppresults1},
and others shown in \myreffig{fig:vid2}.
{
Besides ENet, FRVSR and DUF as baselines, we further compare our TecoGAN model to RBPN~\citep{haris2019recurrent} and EDVR~\citep{wang2019edvr}.
Note that in contrast to TecoGAN with 3 million trainable weights, the latter two use substantially larger networks which have more than 12 and 20 million weights respectively,
and EDVR is trained using bi-cubic down-sampling.
Thus, in the following quantitative and qualitative comparisons, results and metrics for EDVR are calculated using bi-cubic down-sampled images, while other models use LR inputs with a Gaussian blur.
In the supplemental web-page,
{Sec. 2 contains video results for the stills shown in \myreffig{fig:vid2},
while
Sec. 3 shows video comparisons of Vid4 scenes.}
}

Trained with down-sampled inputs with Gaussian blur,
the VSR TecoGAN model can similarly work with original images
that were not down-sampled or filtered,
such as a data-set of real-world photos.
In \myreffig{fig:real}, we compared our results to two other methods~\cite{liao2015video, tao2017spmc} that have used the same dataset.
With the help of adversarial learning, our model is able to generate improved and realistic details in down-sampled images as well as captured images.

For UVT tasks, we train models for Obama and Trump translations,
LR- and HR- smoke simulation translations, as well as translations
between smoke simulations and real-smoke captures.
While smoke simulations usually contain
strong numerical viscosity with details limited by the simulation resolution,
the real smoke from Eckert et al.
\shortcite{eckert2018coupled} contains vivid motions
with many vortices and high-frequency details.
As shown in \myreffig{fig:UVTmore}, our method can be used to
narrow the gap between simulations and real-world phenomena.

While visual results discussed above provide a first indicator of the
quality our approach achieves, quantitative evaluations are crucial for automated
evaluations across larger numbers of samples.
Below we focus more on the VSR task as ground-truth data is available.
We conduct user studies and present evaluations of the different models
w.r.t. established spatial metrics. We also motivate and propose two novel temporal metrics
to quantify temporal coherence.

For evaluating image SR, \citet{blau2018perception}
demonstrated that there is an inherent trade-off between the perceptual quality of the result
and the distortion measured with vector norms or low-level structures such as PSNR and SSIM.
On the other hand, metrics based on deep feature maps such as LPIPS~\cite{zhang2018unreasonable}
can capture more semantic similarities.
We measure the PSNR and LPIPS using the Vid4 scenes.
{With a PSNR decrease of less than 2dB over DUF (which has twice the model size),
the LPIPS score of TecoGAN shows an improvement of more than {40\mypercent{}}. The other baselines are outperformed by similar margins.
Even compared to the large EDVR model using down-sampled inputs without Gaussian blur,
TecoGAN still yields a 30\mypercent{} improvement in terms of LPIPS.}

While traditional temporal metrics based on
vector norm differences of warped frames, e.g. T-diff  {$=\left\| g_t - W( g_{t-1}, v_t )\right\|_1$ }~\cite{chen2017coherent},
can be easily deceived by very blurry results, {e.g. bi-cubic interpolated ones},
we propose to use a tandem of {two new metrics},
tOF and tLP, to measure the consistence over time.
tOF measures the pixel-wise difference of motions estimated from sequences,
and tLP measures perceptual changes over time using deep feature map:
{\small
\vspace{-3pt}
\resizeEq{\text{tOF} &= \left\| OF(b_{t-1}, b_t) - OF(g_{t-1}, g_t) \right\|_1 & \text{ and }\\
\text{\selflp{}} &= \left\| LP(b_{t-1}, b_t) - LP(g_{t-1}, g_t) \right\|_1 \text{.} &
}{eq:tmetrics}{!}}
$OF$ represents an optical flow estimation with the {Farneback}~\shortcite{farneback2003two}
algorithm and $LP$ is the perceptual LPIPS metric.
In tLP, the behavior of the reference is also considered,
as natural videos exhibit a certain degree of change over time.
In conjunction, both pixel-wise differences and perceptual changes
are crucial for quantifying realistic temporal coherence.
While they could be combined into a single score, we list
both measurements separately, as their relative
importance could vary in different application settings.

Our evaluation with these temporal metrics
in \myreftab{tab:metrics} shows that all temporal
adversarial models outperform spatial adversarial ones
and the full TecoGAN model performs very well:
With a large amount of spatial detail, it still achieves good temporal coherence, on par with non-adversarial methods such as DUF, FRVSR, {RBPN and EDVR.
These results are also visualized in \myreffig{fig:metrics}.}
{For VSR,} we have confirmed these automated evaluations with several user studies  (details in \myrefapp{app:userstudy}).
Across all of them, we find that the majority of the participants considered
the TecoGAN results to be closest to the ground truth, when comparing to bi-cubic interpolation, ENet, FRVSR and DUF.

For the UVT tasks, where no ground-truth data is available, we can still evaluate tOF and tLP metrics
by comparing the motion and the perceptual changes of the output data w.r.t. the ones from the input data
{
, i.e., {\footnotesize
$\text{tOF} =\left\| OF(a_{t-1}, a_t) - OF(g^{a\rightarrow b}_{t-1}, g^{a\rightarrow b}_t) \right\|_1$}
and {\footnotesize \selflp{}$ = \left\| LP(a_{t-1}, a_t) - LP(g^{a\rightarrow b}_{t-1}, g^{a\rightarrow b}_t) \right\|_1$}.
With sharp spatial features and coherent motion,
TecoGAN outperforms {CycleGAN and RecycleGAN} on the Obama\&Trump dataset, as shown in \myreftab{tab:UVTmetrics},
although it is worth pointing out that tOF is less informative in this case,
as the motion in the target domain is not necessarily pixel-wise aligned with the input.
{
While RecycleGAN uses an {L2-based cycle} loss that leads to undesirable smoothing,
{Park et al.~\shortcite{park2019preserving} propose to use temporal-cycle losses in together with a VGG-based
content preserving loss (we will refer to this method as {\em STC-V2V} below).}
While the evaluation of temporal metrics for TecoGAN and
STC-V2V is very close, \myreffig{fig:UVTCMP} shows that our results
contain sharper spatial details, such as the eyes and eyebrows of Obama as well as the wrinkles of Trump. This is illustrated in Sec. 2.2 of the supplemental web-page.}
{Overall, TecoGAN successfully generates spatial details, on par with CycleGAN.
TecoGAN also achieves very good tLP scores thanks to the
supervision of temporal coherence, on par with previous work~\cite{bansal2018recycle, park2019preserving},
despite inferring outputs with improved spatial complexity.}

In line with VSR, a perceptual evaluation by humans {in a user study confirms
our metric evaluations for the UVT task. The participants consistently prefer
TecoGAN results over CycleGAN and RecycleGAN.
The corresponding scores are given in the right column of \myreftab{tab:UVTmetrics}.}
}

{\setlength{\tabcolsep}{0.9pt}
\begin{table}[b]\footnotesize
{
\centering
\caption{{For the Obama\&Trump dataset, the averaged tLP and tOF evaluations closely correspond to our
user studies. The table below summarizes user preferences as Bradley-Terry scores.
Details are given in \myrefapp{app:userstudy} and Sec. 3 of the supplemental web-page.
}}
\label{tab:UVTmetrics}
\begin{tabular}{l|c|c|c|c|c|c||c|c}\hline
UVT scenes & \multicolumn{2}{c|}{Trump$\rightarrow$Obama}
& \multicolumn{2}{c|}{Obama$\rightarrow$Trump}
& \multicolumn{2}{c||}{AVG}
& \multicolumn{2}{c}{$^{\text{User Studies}\uparrow,}_\text{ref. to}$} \\[3pt]\hline
metrics    & \multicolumn{1}{c|}{tLP$\downarrow$} & \multicolumn{1}{c|}{tOF$\downarrow$}
& \multicolumn{1}{c|}{tLP$\downarrow$} & \multicolumn{1}{c|}{tOF$\downarrow$}
& \multicolumn{1}{c|}{tLP$\downarrow$} & \multicolumn{1}{c||}{tOF$\downarrow$}
& \multicolumn{1}{c|}{$^\text{original}_\text{input}$} & \multicolumn{1}{c}{
$^\text{arbitrary}_\text{target}$} \\[2pt]\hline
CycleGAN   & 0.0176 & 0.7727 & 0.0277 & 1.1841 & 0.0234 & 0.9784 & 0.0 & 0.0\\\hline
RecycleGAN & \textbf{0.0111} & 0.8705 & 0.0248 & 1.1237 & 0.0179 & 0.9971 & 0.994 & 0.202 \\\hline
STC-V2V
& {0.0143} & {0.7846} & {\textbf{0.0168}} & {0.927} & {\textbf{ 0.0156}} & {0.8561} & - & -\\\hline
TecoGAN    & 0.0120 & \textbf{0.6155} & {0.0191} & \textbf{0.7670} &\textbf{ 0.0156} & \textbf{0.6913} & \textbf{1.817} & \textbf{0.822} \\\hline
\end{tabular}
}
\end{table}
}

\section{Discussion and Limitations}

In paired as well as unpaired data domains,
we have demonstrated that it is possible to learn stable
temporal functions with GANs
thanks to the proposed discriminator architecture and \ppl{} loss.
We have shown that this yields coherent
and sharp details for VSR problems that go beyond what can be achieved
with direct supervision.
In UVT, we have shown that our architecture
guides the training process to successfully establish the spatio-temporal cycle consistency between two domains.
These results are reflected in the proposed metrics and confirmed by user studies.

While our method generates very realistic results for a wide range of natural images, our method can lead to
temporally coherent yet sub-optimal details in
certain cases such as under-resolved faces and text in VSR,
or UVT tasks with strongly different motion between two domains.
For the latter case, it would be interesting to apply both our method and motion translation from concurrent work~\cite{chen2019mocycle}.
This can make it easier for the generator to learn from our temporal self-supervision.
{The proposed temporal self-supervision also has potential to improve other tasks such as
video in-painting and video colorization.
In these multi-modal problems, it is especially important to preserve long-term temporal consistency.}
For our method, the interplay of the different loss terms in the non-linear training
procedure does not provide a guarantee that all goals are fully reached every time.
However, we found our method to be stable over a large number of training runs
and we anticipate that it will provide a very useful basis for
a wide range of generative models for temporal data sets.

\begin{acks}
This work was supported by the ERC Starting Grant realFlow (StG-2015-637014) and the Humboldt Foundation through the Sofja Kovalevskaja Award.
We would like to thank Kiwon Um for helping with the user studies.
\end{acks}

{
\bibliographystyle{ACM-Reference-Format}
\bibliography{egbib}
}

\clearpage
\appendix
\newpage
\setcounter{page}{1}

\begin{strip}
{
\vspace{-36pt}
\section*{\huge Learning Temporal Coherence via Self-Supervision \newline
for GAN-based Video Generation \newline\newline
The Appendix
} 
}
\end{strip}

\noindent
In the following, we first
give details of the proposed temporal evaluation metrics, and present the corresponding quantitative comparison of our method versus a range of others (\myrefapp{app:metrics}).
The user studies we conduected are in support of our
TecoGAN network and proposed temporal metrics, and explained in \myrefapp{app:userstudy}.
Then, we give technical details of our spatio-temporal discriminator (\myrefapp{app:mcinD})
and the proposed PP loss (\myrefapp{app:PPaug}).
Details of network architectures and training parameters are listed (\myrefapp{app:netarc}, \myrefapp{app:training}).
Lastly, we discuss the performance of our approach in \myrefapp{app:performance}.
\newline

\section{Metrics and Quantitative Analysis}
\label{app:metrics}
\subsection{Spatial Metrics}

In order to be able to compare our results with single-image methods,
we evaluate all VSR methods with the purely spatial metrics PSNR
together with the human-calibrated LPIPS metric~\citep{zhang2018unreasonable}.
While higher PSNR values indicate a better pixel-wise accuracy,
lower LPIPS values represent better perceptual quality and closer semantic similarity.
{Note that both metrics are agnostic to changes over time, and hence do not suffice to really evaluate video data.}

Mean values of the Vid4 scenes~\cite{liu2011bayesian}
are shown on the top of \myreftab{tab:vid4metrics}.
Trained with direct vector norms losses, {FRVSR, DUF, EDVR, and RBPN}
achieve high PSNR scores.
However, the undesirable smoothing induced by these losses manifests themselves in larger LPIPS distances.
ENet, on the other hand, with no information from neighboring frames, yields the lowest PSNR
and achieves an LPIPS score that is only slightly better than DUF and FRVSR.
The TecoGAN model with adversarial training achieves an excellent LPIPS score,
with a PSNR	decrease of less than 2dB over DUF.
This is very reasonable, since
PSNR and perceptual quality
were shown to be anti-correlated~\citep{blau2018perception},
especially in regions where PSNR is very high.
Based on good perceptual quality and reasonable pixel-wise accuracy,
TecoGAN outperforms all other methods by more than 30\mypercent{} for LPIPS.

{\setlength{\tabcolsep}{0.3pt}
\begin{table}[t!]
\begin{center}
\footnotesize
\caption{Metrics evaluated for the VSR Vid4 scenes.}\label{tab:vid4metrics}
\vspace{-12pt}
\begin{tabular}{l|c|c|c|c|c|c|c||c|c|c|c}
\hline
PSNR$\uparrow$ & BIC   & ENet  & FRVSR  & DUF    & {RBPN}  & {EDVR}   & TecoGAN & \TecoSmallTwo & DsOnly  & DsDt  & DsDtPP \\\hline
calendar       & 20.27 & 19.85 & 23.86  & 24.07  & 23.88 & 23.97  &  23.21  & 23.35         & 22.23   & 22.76 & 22.95   \\
foliage        & 23.57 & 21.15 & 26.35  & 26.45  & 26.32 & 26.42  &  24.26  & 25.13         & 22.33   & 22.73 & 25.00   \\
city           & 24.82 & 23.36 & 27.71  & 28.25  & 27.51 & 27.76  &  26.78  & 26.94         & 25.86   & 26.52 & 27.03   \\
walk           & 25.84 & 24.90 & 29.56  & 30.58  & 30.59 & 30.92  &  28.11  & 28.14         & 26.49   & 27.37 & 28.14   \\
average        & 23.66 & 22.31 & 26.91  & \textbf{27.38}
& 27.15 & 27.34  &  25.57  & 25.89         & 24.14   & 24.75 & 25.77   \\\hline
${\text{LPIPS}\downarrow}{\times 10}$ &
BIC   & ENet  & FRVSR  & DUF    & {RBPN}  & {EDVR}   & TecoGAN & \TecoSmallTwo & DsOnly  & DsDt  & DsDtPP \\\hline
calendar       & 5.935 & 2.191 & 2.989  & 3.086  & 2.654 & 2.296  &  1.511  & 2.142         & 1.532   & 2.111 & 2.112   \\
foliage        & 5.338 & 2.663 & 3.242  & 3.492  & 3.613 & 3.485  &  1.902  & 1.984         & 2.113   & 2.092 & 1.902   \\
city           & 5.451 & 3.431 & 2.429  & 2.447  & 0.233 & 2.264  &  2.084  & 1.940         & 2.120   & 1.889 & 1.989   \\
walk           & 3.655 & 1.794 & 1.374  & 1.380  & 1.362 & 1.291  &  1.106  & 1.011         & 1.215   & 1.057 & 1.051   \\
average        & 5.036 & 2.458 & 2.506  & 2.607  & 2.511 & 2.356  &  \textbf{1.623}
& 1.743         & 1.727   & 1.770 & 1.733   \\\hline
${\text{\ofmae}\downarrow}{\times 10}$ &
BIC   & ENet  & FRVSR  & DUF    & {RBPN}  & {EDVR}    & TecoGAN & \TecoSmallTwo & DsOnly  & DsDt  & DsDtPP \\\hline
calendar       & 4.956 & 3.450 & 1.537  & 1.134  & 1.068 & 0.986   &  1.342  & 1.403         & 1.609   & 1.683 & 1.583   \\
foliage        & 4.922 & 3.775 & 1.489  & 1.356  & 1.234 & 1.144   &  1.238  & 1.444         & 1.543   & 1.562 & 1.373   \\
city           & 7.967 & 6.225 & 2.992  & 1.724  & 1.584 & 1.446   &  2.612  & 2.905         & 2.920   & 2.936 & 3.062   \\
walk           & 5.150 & 3.203 & 2.569  & 2.127  & 1.994 & 1.871   &  2.571  & 2.765         & 2.745   & 2.796 & 2.649   \\
average        & 5.578 & 4.009 & 2.090  & 1.588  & 1.473 & \textbf{1.367}
&  1.897  & 2.082         & 2.157   & 2.198 & 2.103   \\\hline
${\text{\selflp}\downarrow}{\times 100}$ &
BIC   & ENet  & FRVSR  & DUF    & {RBPN}  & {EDVR}    & TecoGAN & \TecoSmallTwo & DsOnly  & DsDt  & DsDtPP \\\hline
calendar       & 3.258 & 2.957 & 1.067  & 1.603  & 0.802 & 0.622   &  0.165  & 1.087         & 0.872   & 0.764 & 0.670   \\
foliage        & 2.434 & 6.372 & 1.644  & 2.034  & 1.927 & 1.998   &  0.894  & 0.740         & 3.422   & 0.493 & 0.454   \\
city           & 2.193 & 7.953 & 0.752  & 1.399  & 0.432 & 1.060   &  0.974  & 0.347         & 2.660   & 0.490 & 0.140   \\
walk           & 0.851 & 2.729 & 0.286  & 0.307  & 0.271 & 0.171   &  0.653  & 0.635         & 1.596   & 0.697 & 0.613   \\
average        & 2.144 & 4.848 & 0.957  & 1.329  & 0.911 & 0.982   & \textbf{ 0.668}
& 0.718         & 2.160   & 0.614 & 0.489   \\\hline
\end{tabular}\\
\setlength{\tabcolsep}{0.2pt}
\begin{tabular}{l|c|c|c|c|c|c|c|c|c|c}
\hline
${\text{T-diff}\downarrow}{\times 100}$ &
BIC   & ENet  & FRVSR  & DUF    & TecoGAN & \TecoGANsmall & DsOnly  & DsDt  & DsDtPP& GT   \\\hline
calendar       & 2.271 & 9.153 & 3.212  & 2.750  &  4.663  & 3.496         & 6.287   & 4.347 & 4.167  & 6.478 \\
foliage        & 3.745 & 11.997& 3.478  & 3.115  &  5.674  & 4.179         & 8.961   & 6.068 & 4.548  & 4.396 \\
city           & 1.974 & 7.788 & 2.452  & 2.244  &  3.528  & 2.965         & 4.929   & 3.525 & 2.991  & 4.282 \\
walk           & 4.101 & 7.576 & 5.028  & 4.687  &  5.460  & 5.234         & 6.454   & 5.714 & 5.305  & 5.525 \\
average        & 3.152 & 9.281 & 3.648  & 3.298  &  4.961  & 4.076         & 6.852   & 5.071 & 4.369  & 5.184 \\\hline
\end{tabular}
\vspace{-12pt}
\end{center}
\end{table}}
{
\setlength{\tabcolsep}{2.5pt}
\setlength\extrarowheight{-0.8pt}
\begin{table}[h!]\footnotesize
\caption{Metrics evaluated for VSR of ToS scenes.}
\label{tab:ToSmetrics}\vspace{-12pt}
\begin{tabular}{r|c|c|c|c|c|c|c}
\hline
PSNR$\uparrow$
&  BIC   & ENet  & FRVSR  & DUF    & {RBPN}  & {EDVR} & TecoGAN \\\hline
room        & 26.90  & 25.22 & 29.80  & 30.85  & 26.82 & 31.12 &  29.31   \\
bridge      & 28.34  & 26.40 & 32.56  & 33.02  & 28.56 & 32.88 &  30.81   \\
face        & 33.75  & 32.17 & 39.94  & 40.23  & 33.74 & 41.57 &  38.60   \\
average     & 29.58  & 27.82 & 34.04  & 34.60  & 29.71 & \textbf{35.02} &  32.75   \\\hline
${\text{LPIPS}\downarrow}{\times 10}$
&  BIC  & ENet  & FRVSR  & DUF    & {RBPN} & {EDVR} & TecoGAN  \\\hline
room        & 5.167 & 2.427 & 1.917  & 1.987  & 2.054 & 2.232 &  1.358  \\
bridge      & 4.897 & 2.807 & 1.761  & 1.684  & 1.845 & 1.663 &  1.263  \\
face        & 2.241 & 1.784 & 0.586  & 0.517  & 0.728 & 0.613 &  0.590  \\
average     & 4.169 & 2.395 & 1.449  & 1.414  & 1.565 & 1.501 &  \textbf{1.086}  \\\hline
${\text{\ofmae}\downarrow}{\times 10}$
&  BIC  & ENet  & FRVSR  & DUF     & {RBPN} & {EDVR} & TecoGAN \\\hline
room        & 1.735 & 1.625 & 0.861  & 0.901   & 0.821 & 0.769 &  0.737  \\
bridge      & 5.485 & 4.037 & 1.614  & 1.348   & 1.354 & 1.140 &  1.492  \\
face        & 4.302 & 2.255 & 1.782  & 1.577   & 1.612 & 1.383 &  1.667  \\
average     & 4.110 & 2.845 & 1.460  & 1.296   & 1.287 & \textbf{1.113} &  1.340  \\\hline
${\text{\selflp}\downarrow}{\times 100}$
&  BIC  & ENet  & FRVSR  & DUF    & {RBPN}  & {EDVR} & TecoGAN \\\hline
room        & 1.320 & 2.491 & 0.366  & 0.307  & 0.263 & 0.252 &  0.590  \\
bridge      & 2.237 & 6.241 & 0.821  & 0.526  & 0.821 & 1.030 &  0.912  \\
face        & 1.270 & 1.613 & 0.290  & 0.314  & 0.364 & 0.396 &  0.379  \\
average     & 1.696 & 3.827 & 0.537  & \textbf{0.403}  & 0.531 & 0.628 &  0.664  \\\hline
\end{tabular}\vspace{-6pt}
\end{table}}

\begin{figure*}[t]\footnotesize \centering
\begin{overpic}[width=0.4\linewidth]{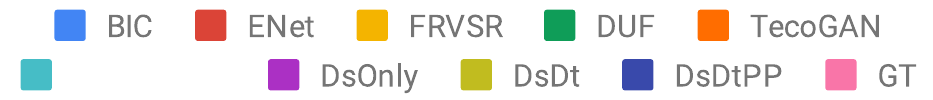}
\put(7.5,1.2){\footnotesize \color{darkgrey}\textsf{{\textbf{\TecoGANsmall}}}}
\end{overpic}\\
\begin{minipage}{0.9\textwidth}\centering
\begin{tabular}{ccc}
\includegraphics[width=0.3\textwidth]{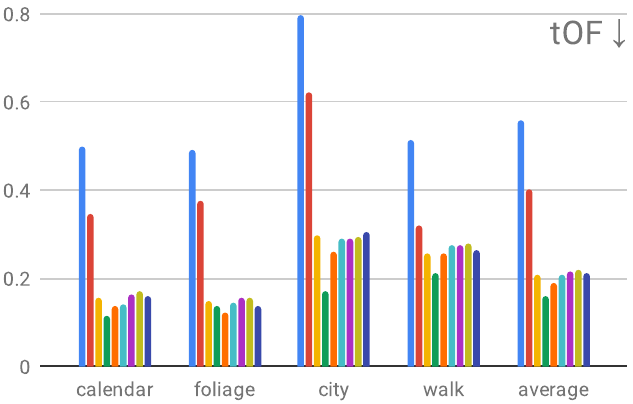} &
\includegraphics[width=0.3\textwidth]{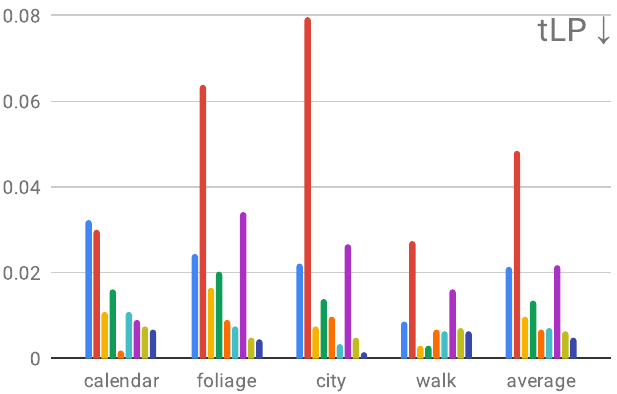} &
\includegraphics[width=0.3\textwidth]{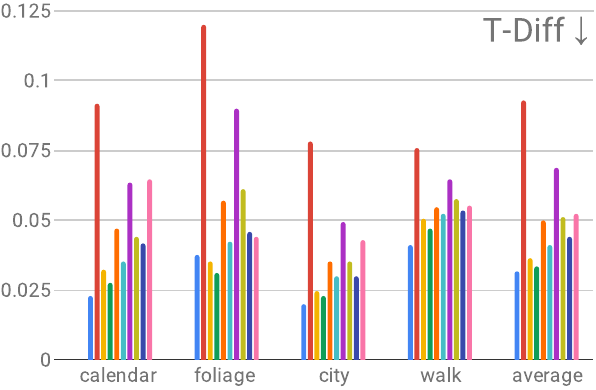}
\end{tabular}
\caption{Bar graphs of temporal metrics for Vid4.}\label{fig:suppgraphs2}
\end{minipage}\\
\begin{minipage}{.3\textwidth} \centering
\includegraphics[width=\linewidth]{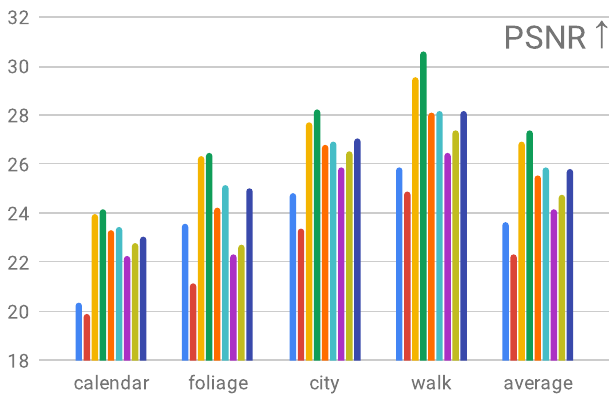}\\
\includegraphics[width=\linewidth]{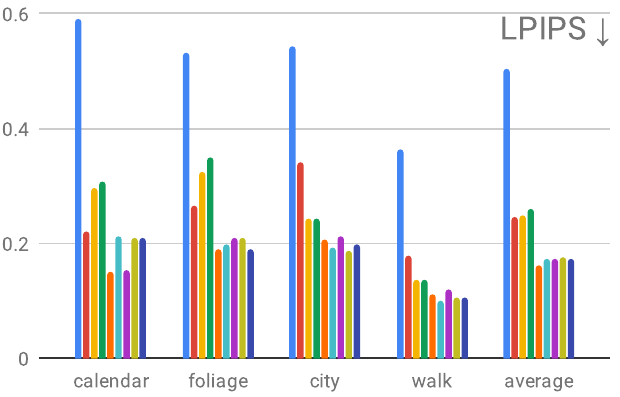}\\
\caption{\footnotesize Spatial metrics for Vid4.}
\label{fig:suppgraphs1}
\end{minipage}
\begin{minipage}{0.6\textwidth} \centering
\begin{tabular}{cc}
\includegraphics[width=0.49\textwidth]{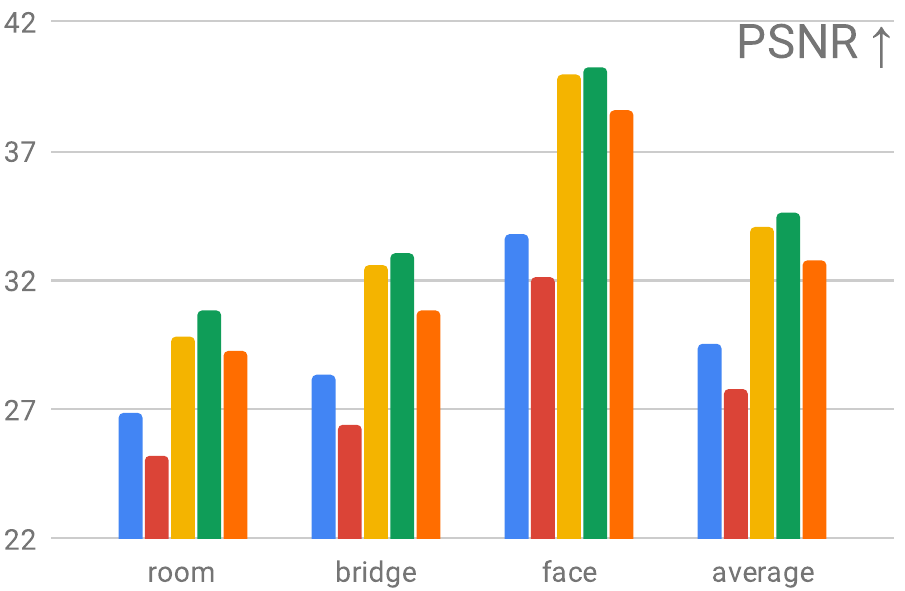}
&\includegraphics[width=0.49\textwidth]{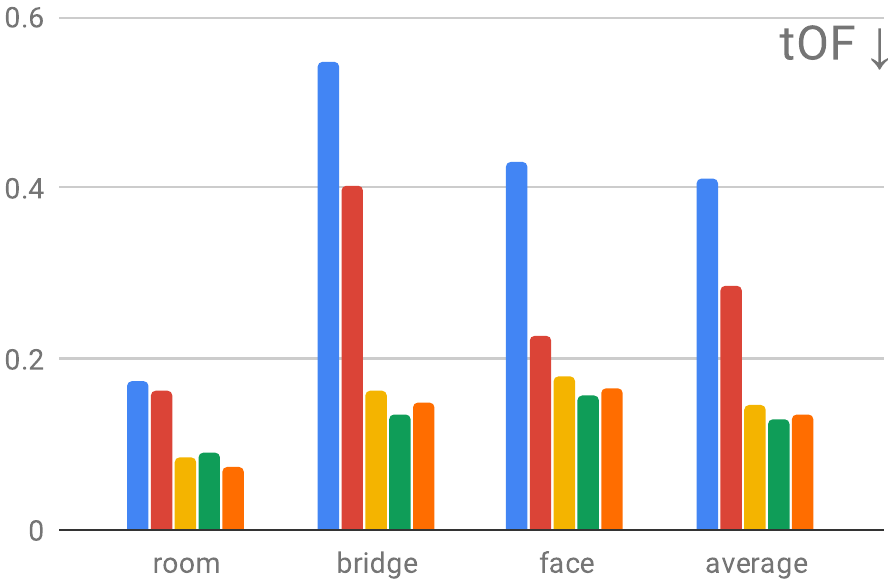}\\ \includegraphics[width=0.49\textwidth]{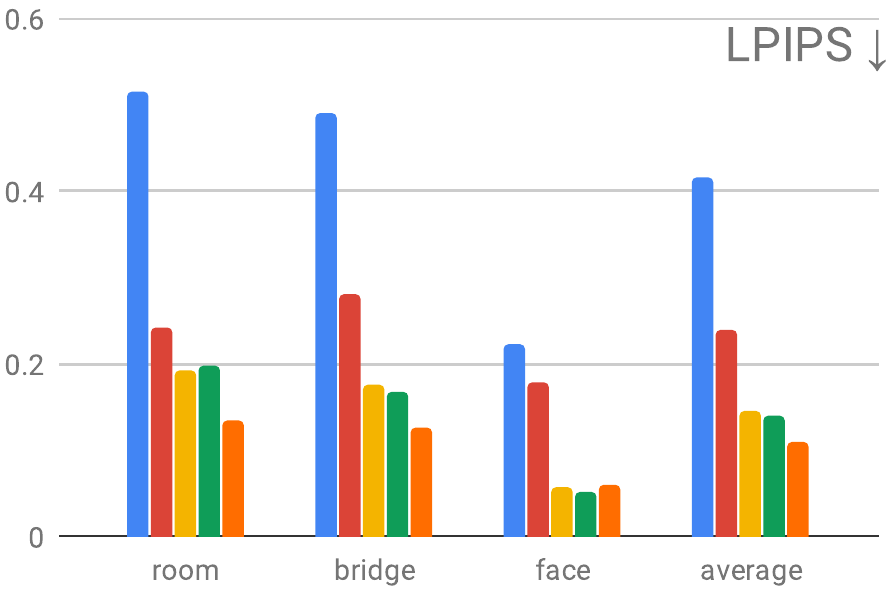}
&\includegraphics[width=0.49\textwidth]{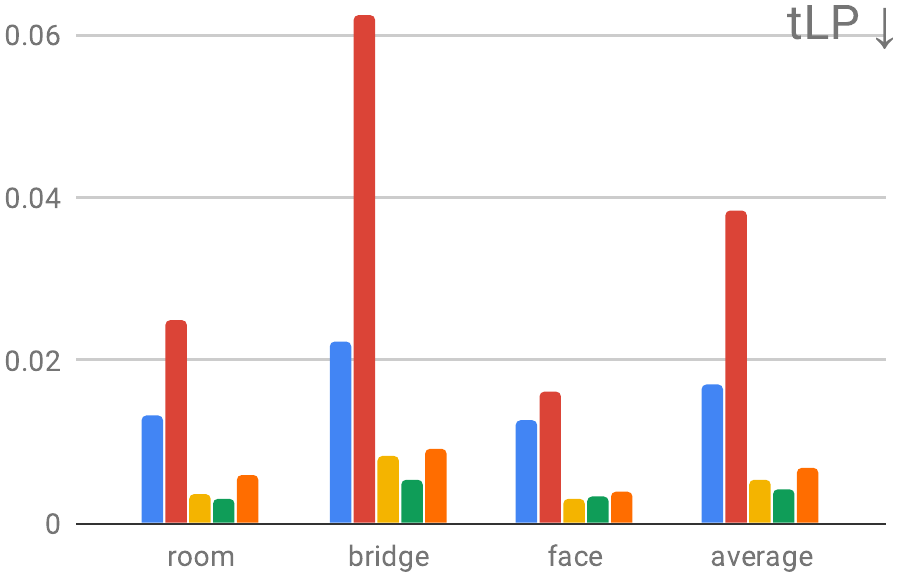}\\
\end{tabular}
\vspace{-6pt}
\caption{Metrics for ToS.}
\label{fig:suppgraphs3}
\end{minipage}
\end{figure*}

\subsection{Temporal Metrics}
For both VSR and UVT,
evaluating temporal coherence
without ground-truth motion
is very challenging.
$L^2$-based temporal metrics such as {\em T-diff}
$= \| g_t - $ $W( g_{t-1}, v_t )\|_1$
was used~\cite{chen2017coherent}
as a rough assessment of temporal differences, and we give corresponding numbers for comparison.
As shown at the bottom of \myreftab{tab:vid4metrics},
T-diff, due to its local nature, is easily deceived by blurry method such as the bi-cubic interrelation
and can not correlate well with visual assessments of coherence.

{By using the proposed metrics, i.e. measuring the pixel-wise motion differences using \text{tOF}
together with the perceptual changes over time using \selflp{},
a more nuanced evaluation can be achieved, as
shown for the VSR task}
in the middle of \myreftab{tab:vid4metrics}.
Not surprisingly, the results of ENet show larger errors for all metrics due to their strongly flickering content.
Bi-cubic up-sampling, DUF, and FRVSR achieve very low T-diff errors due to their smooth results, representing
an easy, but undesirable avenue for achieving coherency. However, the overly smooth
changes of the former two are identified by the \selflp{} scores.
While our DsOnly model generates sharper results at the expense of temporal coherence,
it still outperforms ENet there.
By adding temporal information to discriminators, our DsDt, DsDt+PP, \TecoGANsmall and TecoGAN
improve in terms of temporal metrics. Especially the full TecoGAN model stands out here.
{
For the UVT tasks, temporal motions are evaluated by comparing to the input sequence.
With sharp spatial features and coherent motion,
TecoGAN outperforms previous work on the Obama\&Trump dataset,
as shown in \myreftab{tab:UVTmetrics}.}

\subsection{Spatio-temporal Evaluations}
\setlength{\columnsep}{6pt}
\begin{figure}
\centering
\includegraphics[width=\linewidth]{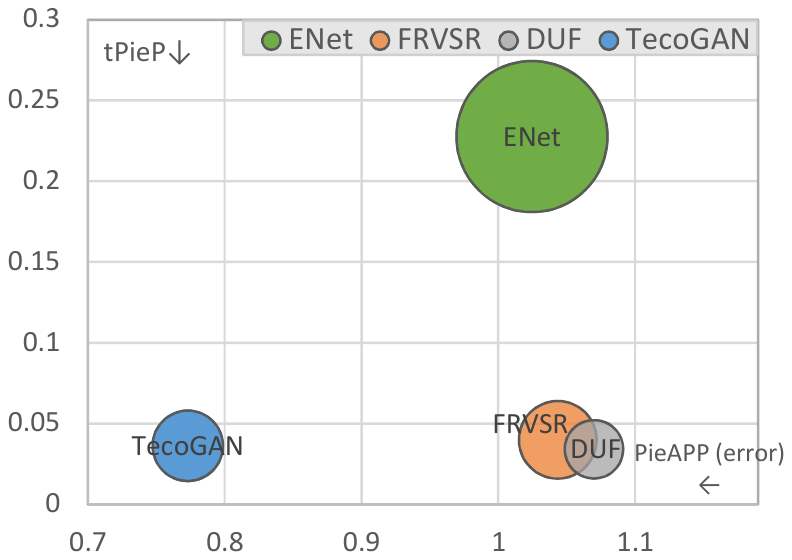}\\\vspace{4pt}
{\setlength{\tabcolsep}{6pt}
\centering
\begin{tabular}{|l|l|l|l|l|l|}
\hline
{\em \selfpieapp{}}${\downarrow}$& BIC & ENet & FRVSR & DUF & TecoGAN \\ \hline
calendar & 0.091   & 0.194   & 0.023    & 0.028  & \textbf{0.021}      \\ \hline
foliage  & 0.155    & 0.276   & 0.040    & 0.037  & \textbf{0.036}      \\ \hline
city     & 0.136   & 0.286   & \textbf{0.025}    & 0.0283  & 0.0276      \\ \hline
walk     & 0.064   & 0.155    & 0.072    & \textbf{0.042}  & 0.060       \\ \hline
\end{tabular}
}
\caption{Tables and visualization of perceptual metrics
computed with PieAPP \citep{prashnani2018pieapp}
(instead of LPIPS used in \myreffig{fig:metrics} previously) on ENet, FRVSR, DUF and TecoGAN for the VSR of Vid4.
Bubble size indicates the \ofmae{} score.}
\label{fig:tPieAPPcomp}
\end{figure}
Since temporal metrics can trivially be reduced for blurry
image content, we found it important to evaluate results
with a combination of spatial and temporal metrics.
Given that perceptual metrics are already widely used for image
evaluations, we believe it is the right time to consider perceptual changes in temporal evaluations,
as we did with our proposed temporal coherence metrics.
{Although not perfect, they are not as easily deceived
as simpler metrics}.
Specifically, tOF is more robust than a direct pixel-wise metric as it compares motions instead of image content.
In the supplemental material (Sec. 6), we visualize the motion difference and it can well reflect the visual inconsistencies.
On the other hand, we found that
our formulation of {\selflp{}} is a general concept that
can work reliably with different perceptual metrics:
When repeating the tLP evaluation with the PieAPP metric~\citep{prashnani2018pieapp}
instead of $LP$, i.e.,
{\footnotesize {\em \selfpieapp{}} = $\left\| f(b_{t-1}, b_t) - f(g_{t-1}, g_t) \right\|_1$},
where f($\cdot$) indicates the perceptual error function of PieAPP,
we get close to identical results, as shown in
~\myreffig{fig:tPieAPPcomp}.
The conclusions from {\em \selfpieapp{}} also closely match
the LPIPS-based evaluation: our network architecture can generate
realistic and temporally coherent detail, and
the metrics we propose allow for a stable,
automated evaluation of the temporal perception of a generated video sequence.

Besides the previously evaluated the Vid4 dataset, with
graphs shown in \myreffig{fig:suppgraphs2}, \ref{fig:suppgraphs1},
we also get similar evaluation results on the {\em Tears of Steel} data-sets (room, bridge, and face, in the following referred to as {\em ToS} scenes)
and corresponding results are shown in \myreftab{tab:ToSmetrics} and \myreffig{fig:suppgraphs3}.
In all tests, we follow the procedures of previous work~\citep{jo2018deep,sajjadi2018FRVSR}
to make the outputs of all methods comparable, i.e.,
for all result images, we first exclude spatial borders with a distance of 8 pixels to the image sides,
then further shrink borders such that the LR input image is divisible by 8
and for spatial metrics, we ignore the first two and the last two frames,
while for temporal metrics, we ignore first three and last two frames,
as an additional previous frame is required for inference.
Below, we additionally show user study results for the Vid4 scenes.
By comparing the user study results and
the metric breakdowns shown in \myreftab{tab:vid4metrics},
we found our metrics to reliably capture the human temporal perception.

\begin{figure}[t]\scriptsize 
\centering
\begin{minipage}{.46\textwidth}
\centering
\includegraphics[width=0.99\linewidth]{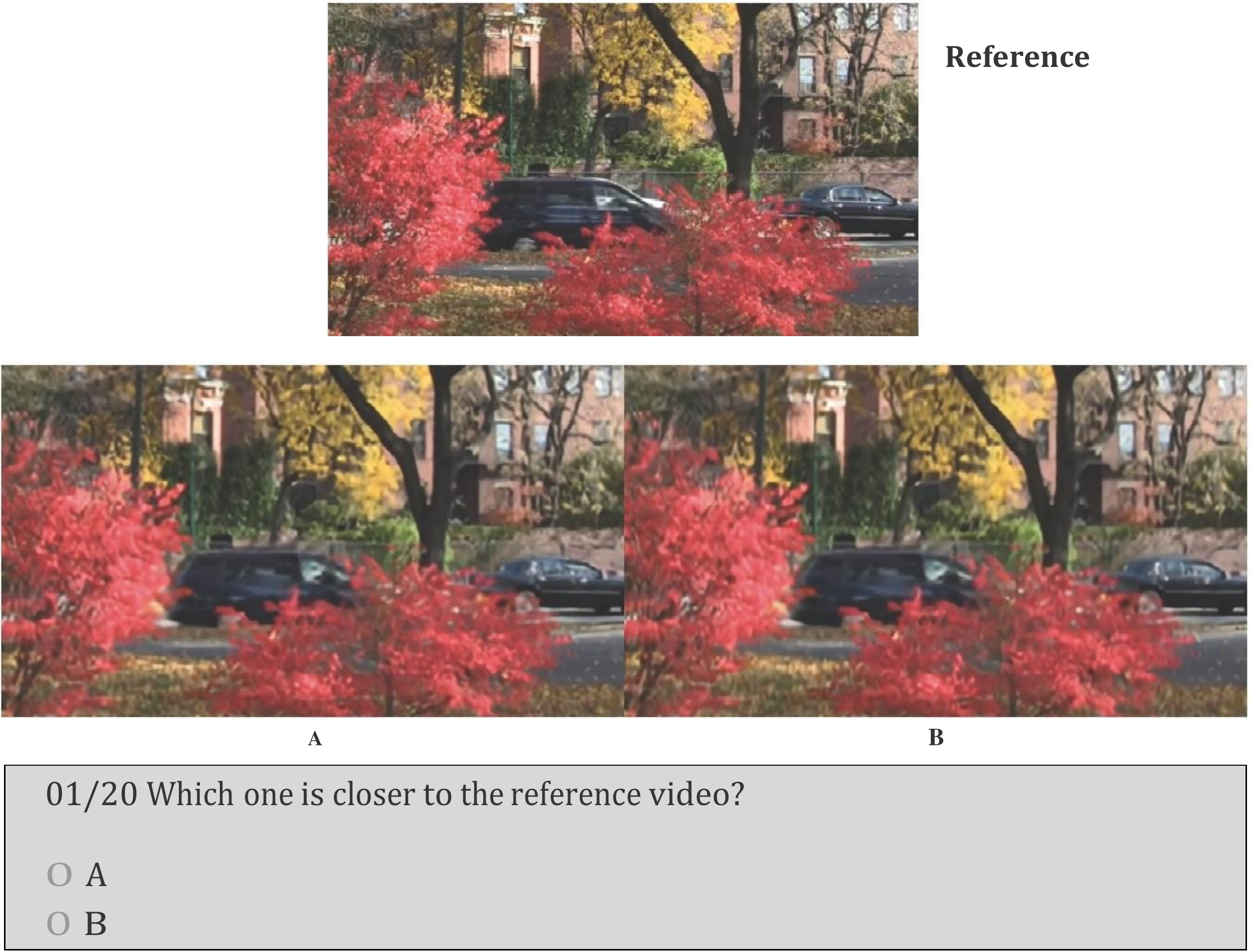}\\
\caption{A sample setup of user study.}
\label{fig:userstudysample}
\end{minipage}
\\
\begin{minipage}{.4\textwidth}
\centering 
\setlength{\tabcolsep}{2pt}
\begin{tabular}{c|rl|rl|rl|rl}
\hline
\multirow{ 2}{*}{Methods}  & \multicolumn{8}{c}{The Bradley-Terry scores (standard error) } \\
\cline{2-9}
& \multicolumn{2}{c|}{calendar} & \multicolumn{2}{c|}{foliage}
& \multicolumn{2}{c|}{city}& \multicolumn{2}{c}{walk} \\
\hline
Bi-cubic   & 0.000  & (0.000) & 0.000 & (0.000) & 0.000 & (0.000) & 0.000 & (0.000) \\\hline
ENet        & 1.834  & (0.228) & 1.634 & (0.180) & 1.282 & (0.205) & 1.773 & (0.197) \\
FRVSR       & 3.043  & (0.246) & 2.177 & (0.186) & 3.173 & (0.240) & 2.424 & (0.204) \\
DUF         & 3.468  & (0.252) & 2.243 & (0.186) & 3.302 & (0.242) & \textbf{3.175} & (0.214) \\
TecoGAN     & \textbf{4.091}  & (0.262) & \textbf{2.769} & (0.194) & \textbf{4.052} & (0.255) & 2.693 & (0.207) \\
\hline
\end{tabular}
\includegraphics[width=0.98\linewidth]{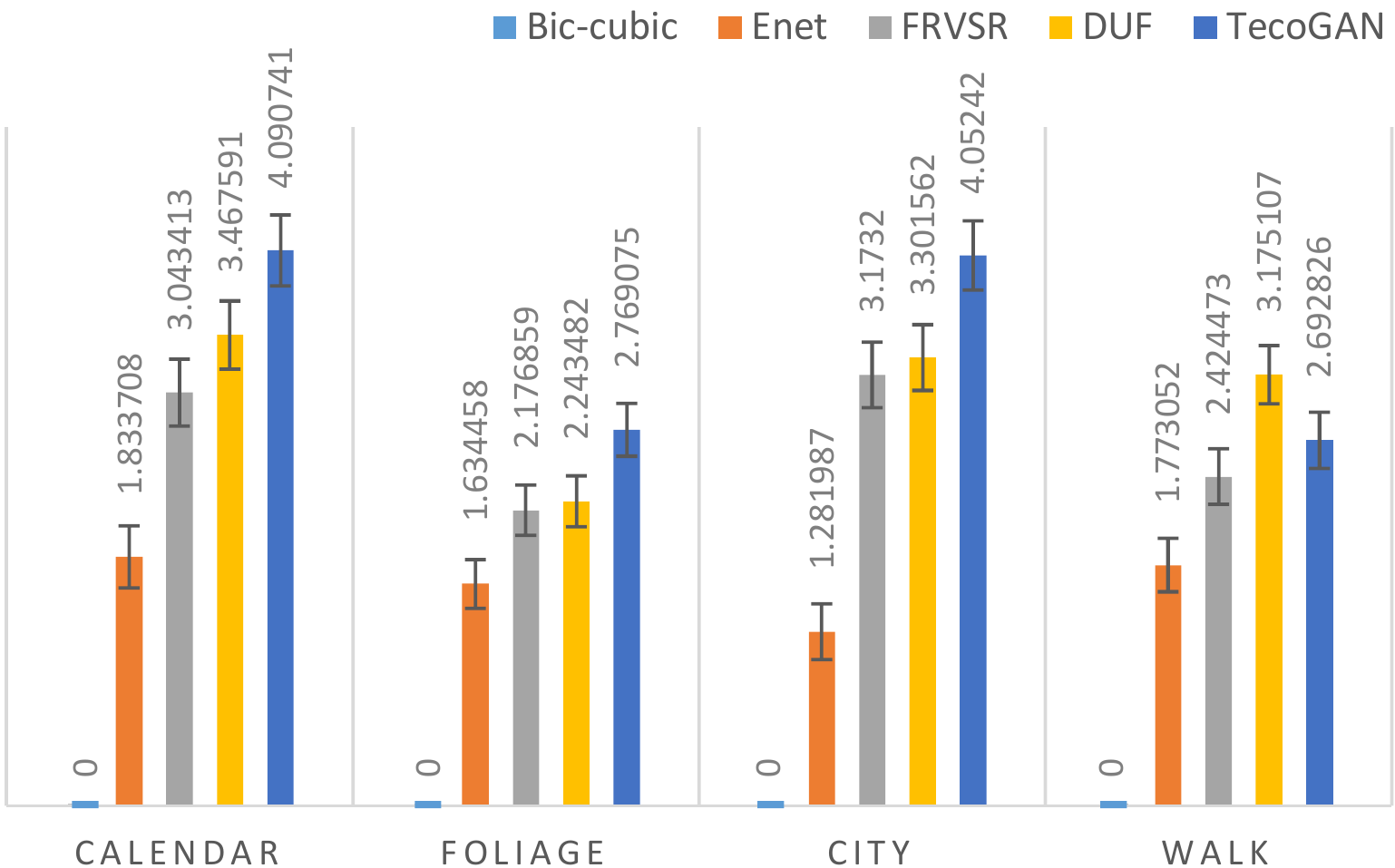}\\
\caption{
{\footnotesize
Tables and bar graphs of Bradley-Terry scores and standard errors for Vid4 VSR.}}
\label{fig:userstudy}
\end{minipage}\\
\end{figure}
\begin{figure*}[t] \footnotesize 

\centering
\setlength{\tabcolsep}{1.6pt}
\begin{minipage}{0.33\textwidth}{
\begin{tabular}{c|c|c|c|c|c|}
\hline
\multicolumn{1}{c|}{\begin{tabular}[c]{@{}c@{}}X-axis: \\
$^\text{Generated Obama}_\text{vs. Obama}$ \end{tabular}} &
\multicolumn{1}{c|}{\begin{tabular}[c]{@{}c@{}}$^\text{Bradley}_\text{Terry}$ \\ scores\end{tabular}} &
\multicolumn{1}{c|}{\begin{tabular}[c]{@{}c@{}}Std. \\ Error\end{tabular}} &
\multicolumn{1}{c|}{\begin{tabular}[c]{@{}c@{}}Y-axis:  \\
$^\text{Generated Obama}_\text{vs. Trump}$ \end{tabular}} &
\multicolumn{1}{c|}{\begin{tabular}[c]{@{}c@{}}$^\text{Bradley}_\text{Terry}$ \\ scores\end{tabular}} &
\multicolumn{1}{c|}{\begin{tabular}[c]{@{}c@{}}Std. \\ Error\end{tabular}}
\\\hline
CycleGAN                           & 0                      & 0
& CycleGAN                           & 0                    & 0 \\ \hline
RecycleGAN                         & 1.322                  & 0.197
&RecycleGAN                         & 1.410                 & 0.208\\ \hline
TecoGAN                            & \textbf{1.520}                  & 0.201
&TecoGAN                            & \textbf{1.958}                 & 0.222\\ \hline
\end{tabular}}
\end{minipage}
\begin{minipage}{0.33\textwidth}{
\begin{tabular}{|c|c|c|c|c|c|}
\hline
\multicolumn{1}{|c|}{\begin{tabular}[c]{@{}c@{}}X-axis: \\
$^\text{Generated Trump}_\text{vs. Trump}$ \end{tabular}} &
\multicolumn{1}{c|}{\begin{tabular}[c]{@{}c@{}}$^\text{Bradley}_\text{Terry}$ \\ scores\end{tabular}} &
\multicolumn{1}{c|}{\begin{tabular}[c]{@{}c@{}}Std. \\ Error\end{tabular}} &
\multicolumn{1}{c|}{\begin{tabular}[c]{@{}c@{}}X-axis: \\
$^\text{Generated Trump}_\text{vs. Obama}$ \end{tabular}} &
\multicolumn{1}{c|}{\begin{tabular}[c]{@{}c@{}}$^\text{Bradley}_\text{Terry}$ \\ scores\end{tabular}} &
\multicolumn{1}{c|}{\begin{tabular}[c]{@{}c@{}}Std. \\ Error\end{tabular}}
\\ \hline
CycleGAN                           & 0.806                 & 0.177
&CycleGAN                           & 0                     & 0                        \\ \hline
RecycleGAN                         & 0                     & 0
&RecycleGAN                         & 0.623                 & 0.182                    \\ \hline
TecoGAN                            & \textbf{1.727}                 & 0.208
&TecoGAN                            & \textbf{1.092}                & 0.182                    \\ \hline
\end{tabular}}
\end{minipage}
\begin{minipage}{0.33\textwidth}{
\begin{tabular}{|c|c|c|c|c|c}\hline
\multicolumn{1}{|c|}{\begin{tabular}[c]{@{}c@{}}X-axis: \\
$^\text{Generated vs.}_\text{Arbitrary Target}$ \end{tabular}} &
\multicolumn{1}{c|}{\begin{tabular}[c]{@{}c@{}}$^\text{Bradley}_\text{Terry}$ \\ scores\end{tabular}} &
\multicolumn{1}{c|}{\begin{tabular}[c]{@{}c@{}}Std. \\ Error\end{tabular}} & \multicolumn{1}{c|}{\begin{tabular}[c]{@{}c@{}}Y-axis:  \\
$^\text{Generated}_\text{vs. the Input}$ \end{tabular}} &
\multicolumn{1}{c|}{\begin{tabular}[c]{@{}c@{}}$^\text{Bradley}_\text{Terry}$ \\ scores\end{tabular}} &
\multicolumn{1}{c}{\begin{tabular}[c]{@{}c@{}}Std. \\ Error\end{tabular}}
\\ \hline
CycleGAN                           & 0                     & 0                        &
CycleGAN                           & 0                     & 0                        \\ \hline
RecycleGAN                         & 0.202                 & 0.118                    &
RecycleGAN                         & 0.994                 & 0.135                    \\ \hline
TecoGAN                            & \textbf{0.822}                 & 0.123                    &
TecoGAN                            & \textbf{1.817}                 & 0.150                    \\ \hline
\end{tabular}}
\end{minipage}\\
\begin{minipage}{\textwidth}
\centering
\includegraphics[width=\linewidth]{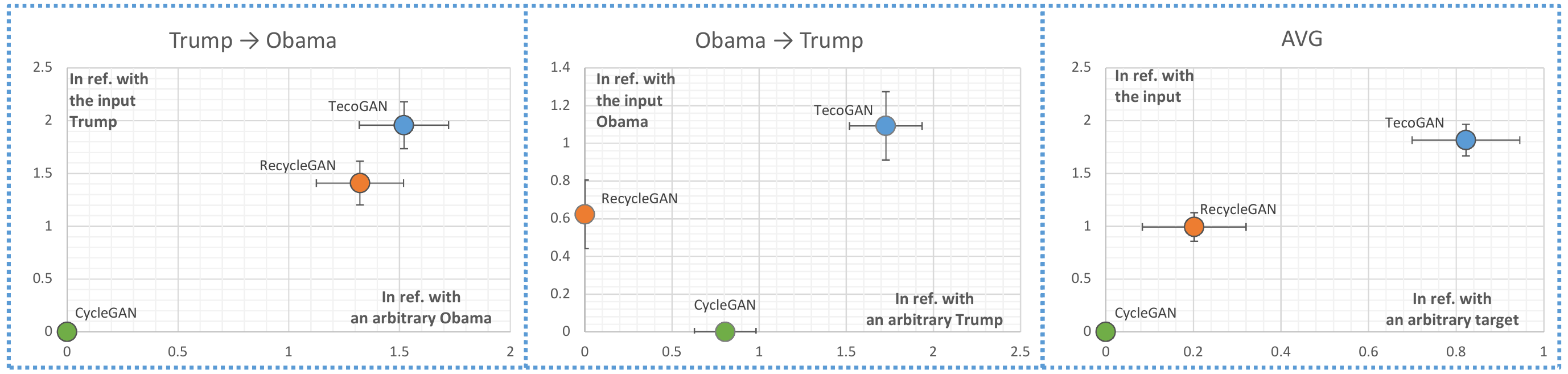}\\
\caption{
{\footnotesize
Tables and graphs of Bradley-Terry scores and standard errors for Obama\&Trump UVT.}}
\label{fig:uvtuser}
\end{minipage}
\end{figure*}
\section{User Studies}
\label{app:userstudy}

We conducted several user studies for the VSR task comparing five different methods:
bi-cubic interpolation, ENet, FRVSR, DUF and  TecoGAN.
The established 2AFC design~\citep{fechner1889elemente,um2017perceptual} is applied,
i.e., participants have a pair-wise choice, with the ground-truth video shown as reference.
One example setup can be seen in \myreffig{fig:userstudysample}.
The videos are synchronized and looped until participants make the final decision.
With no control to stop videos, users
cannot stop or influence the playback,
and hence can focus more on the whole video, instead of specific spatial details.
Videos positions (left/A or right/B) are randomized.

After collecting 1000 votes from 50 users for every scene,
i.e. twice for all possible pairs ($5 \times 4 / 2 = 10$ pairs),
we follow common procedure and compute scores for all models with the Bradley-Terry model~(\citeyear{bradley1952rank}).
The outcomes for the Vid4 scenes can be seen in \myreffig{fig:userstudy}
{(overall scores are listed in \myreftab{tab:metrics} of the main document).}

From the Bradley-Terry scores for the Vid4 scenes
we can see that the TecoGAN model performs very well, and
achieves the first place in three cases, as well as a second place in the walk scene.
{The latter is most likely caused by the overall slightly smoother images
of the walk scene, in conjunction with the presence of several human faces,
where our model can lead to the generation of unexpected details.}
However, overall the user study shows that users preferred the TecoGAN
output over the other two deep-learning methods with a 63.5\mypercent{} probability.

This result also matches with our metric evaluations.
In \myreftab{tab:vid4metrics}, while TecoGAN achieves spatial (LPIPS)
improvements in all scenes,
DUF and FRVSR are not far behind in the walk scene.
In terms of temporal metrics \ofmae{} and \selflp{}, TecoGAN achieves similar
or lower scores compared to FRVSR and DUF for calendar, foliage and city scenes.
The lower performance of our model for the walk scene is likewise
captured by higher \ofmae{}  and \selflp{} scores.
Overall, the metrics confirm the performance of our TecoGAN approach
and match the results of the user studies, which
indicate that our proposed temporal metrics
successfully capture important temporal aspects of human perception.

For UVT tasks which have no ground-truth data,
we carried out two sets of user studies:
One uses an arbitrary sample from the target domain as the reference
and the other uses the actual input from the source domain as the reference.
On the Obama\&Trump data-sets,
we evaluate results from CycleGAN, RecycleGAN, and TecoGAN
following the same modality, i.e. a 2AFC design with 50 users for each run.
E.g., on the left of \myreffig{fig:uvtuser},
users evaluate the generated Obama in reference with the input Trump on the y-axis,
while an arbitrary Obama video is shown as the reference on the x-axis.
Ultimately, the y-axis is more important than the x-axis as it indicates whether
the translated result preserves the original expression.
A consistent ranking of TecoGAN \textgreater~ RecycleGAN \textgreater~ CycleGAN
is shown on the y-axis with clear separations,
i.e. standard errors don't overlap.
The x-axis indicates whether the inferred result matches the general
spatio-temporal content of the target domain. Our TecoGAN
model also receives the highest scores here, although the responses are slightly more spread out.
On the right of \myreffig{fig:uvtuser},
we summarize both studies in a single graph highlighting that the TecoGAN
model is consistently preferred by the participants of our user studies.

\setlength{\columnsep}{6pt}
\begin{figure}[b]
\begin{center}
\begin{overpic}[width=0.86\linewidth]{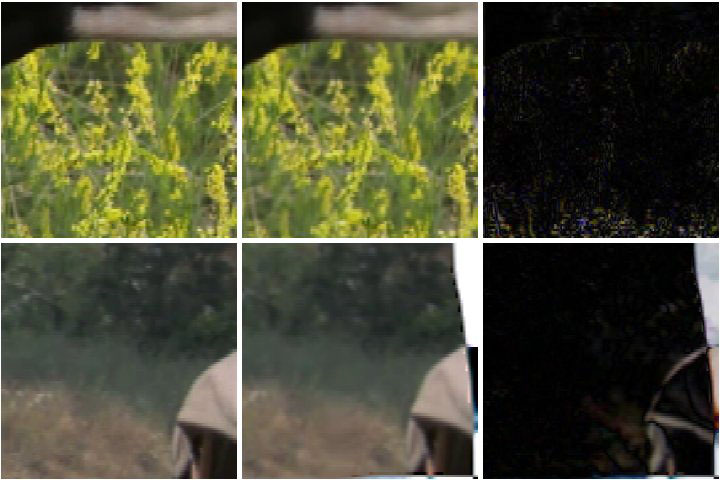}
\put(13.0,67.5){\scriptsize{ \color{black}{$b_{t-1}$}}}
\put(40.0,67.5){\scriptsize{ \color{black}{$W(b_{t-1}, v_t)$}}}
\put(68.0,67.5){\scriptsize{ \color{black}{$\| W(b_{t-1}, v_t) - b_{t}\|_1$}}}
\end{overpic}
\end{center}
\caption{{
Near image boundaries, flow estimation is less accurate
and warping often fails to align content.
The first two columns show original and warped frames,
the third one shows differences after warping (ideally all black).
{The top row shows that structures moving into the view can cause problems, visible at the bottom of the images.
The second row has objects moving out of the view.} } }
\label{fig:warpcrop}
\end{figure}

\section{Technical Details of the \newline Spatio-Temporal Discriminator}
\label{app:mcinD}
\subsection{Motion Compensation Used in Warped Triplet}
In the TecoGAN architecture, $D_{s,t}$ detects the temporal relationships between $I_{wg}$ and $I_{wb}$ with the help of the flow estimation network F. However, at the boundary of images, the output of F is usually less accurate due to the lack of reliable neighborhood information. There is a higher chance that objects move into the field of view, or leave suddenly, which significantly affects the images warped with the inferred motion. An example is shown in \myreffig{fig:warpcrop}.
This increases the difficulty for $D_{s,t}$, as it cannot fully rely on the images being aligned via warping.
To alleviate this problem, we only use the center region
of $I_{wg}$ and $I_{wb}$
as the discriminator inputs and we reset a boundary of 16 pixels.
Thus, for an input resolution of $I_{wg}$ and $I_{wb}$ of $128\times128$ for the VSR task,
the inner part in size of $96\times96$ is left untouched,
while the border regions are overwritten with zeros.

The flow estimation network F
with the loss $\mathcal{L}_{G,F}$ should only be trained to support G in
reaching the output quality as determined by $D_{s,t}$,
but not the other way around.
The latter could lead to F networks that
confuse $D_{s,t}$ with strong distortions of $I_{wg}$ and $I_{wb}$.
In order to avoid the this undesirable case, we stop the gradient back propagation from
$I_{wg}$ and $I_{wb}$ to F. In this way, gradients from $D_{s,t}$ to F are only
back propagated through the generated samples $g_{t-1}, g_t$ and $g_{t+1}$
into the generator network.
As such, $D_{s,t}$ can guide G to improve the image content,
and F learns to warp the previous frame in accordance with the
detail that G can synthesize. However, F
does not adjust the motion estimation only to reduce the adversarial loss.

\subsection{Curriculum Learning for UVT Discriminators}
As mentioned in the main document,
we train the UVT $D_{s,t}$ with 100\mypercent{} spatial triplets at the very beginning.
During training,
25\mypercent{} of them gradually transition to warped triplets
and another 25\mypercent{} transition to original triplets.
{The transitions of the warped triplets are computed with linear interpolation}:
$(1-\alpha) \text{I}_{cg} + \alpha \text{I}_{wg} $, with $\alpha$ growing form 0 to 1.
For the original triplets, we additionally fade the ``warping'' operation out by using
$(1-\alpha) \text{I}_{cg} + \alpha \{ W( g_{t-1}, v_t * \beta ), g_{t}, W( g_{t+1}, v'_t * \beta)\}$,
again with $\alpha$ growing form 0 to 1 and $\beta$ decreasing from 1 to 0.
We found this smooth transition to be helpful for a stable training run.

\section{Data Augmentation and Temporal Constrains in the PP loss}
\label{app:PPaug}

Since training with sequences of arbitrary length
is not possible with current hardware,
problems such as the ``streaking artifacts''
discussed above generally arise for recurrent models.
In the proposed PP loss, both the Ping-Pong data augmentation
and the temporal consistency constraint
contribute to solving these problems.
In order to show their separated contributions, we trained another TecoGAN variant
that only employs the data augmentation without the constraint (i.e., $\lambda_p=0$ in
\myreftab{tab:loss}). Denoted as ``PP-Augment'', we show its results in comparison with
the DsDt and \TecoGANsmall models in \myreffig{fig:PPsep}.
Video results are shown in the in the supplemental material (Sec. 4.5.).

\setlength{\tabcolsep}{1 pt}
\begin{figure}
\centering 
\begin{tabular}{c|c|c}\hline
{DsDt} &
{PP-Augment} & \TecoGANsmall \\ [-1pt]\hline 
{\includegraphics[width=0.32\linewidth, trim={0 12pt 0 30pt},clip]{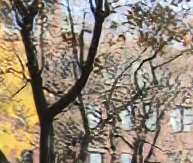}} &
{\includegraphics[width=0.32\linewidth, trim={0 12pt 0 30pt},clip]{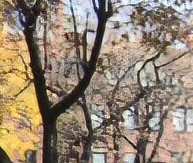}}  &
{\includegraphics[width=0.32\linewidth, trim={0 12pt 0 30pt},clip]{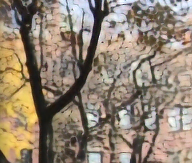}} \\[-2pt]
{\includegraphics[width=0.32\linewidth, trim={0 12pt 0 30pt},clip]{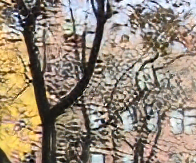}} &
{\includegraphics[width=0.32\linewidth, trim={0 12pt 0 30pt},clip]{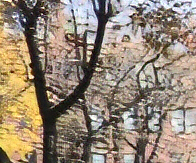}}  &
{\includegraphics[width=0.32\linewidth, trim={0 12pt 0 30pt},clip]{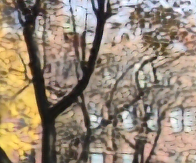}} \\[-2pt]\hline
\end{tabular} \vspace{-8pt}
\caption{
1st \& 2nd row: Frame 15 \& 40 of the {\em Foliage} scene.
While DsDt leads to strong recurrent artifacts early on,
PP-Augment shows similar artifacts later in time (2nd row, middle).
\TecoGANsmall{} model successfully removes these artifacts.}
\label{fig:PPsep}
\end{figure}

During training, the generator of DsDt receives 10 frames, and generators of PP-Augment and \TecoGANsmall{} see 19 frames. While DsDt shows strong recurrent accumulation artifacts early on, the PP-Augment version slightly reduces the artifacts. In \myreffig{fig:PPsep}, it works well for frame 15, but shows artifacts from frame 32 on. Only our regular model (\TecoGANsmall) successfully avoids temporal accumulation for all 40 frames. Hence, with the PP constraint, the model avoids recurrent accumulation of artifacts and works well for sequences that are substantially longer than the training length. Among others, we have tested our model with ToS sequences of lengths 150, 166 and 233. For all of these sequences, the TecoGAN model successfully avoids temporal accumulation or streaking artifacts.

\section{Network Architecture}
\label{app:netarc}

\allowdisplaybreaks
In this section, we use the following notation to specify all network architectures used:
conc$()$ represents the concatenation of two tensors along the channel dimension;
$C/CT($input, kernel\_size, output\_channel, stride\_size$)$ stands for the convolution and transposed convolution operation, respectively;
``+'' denotes element-wise addition;
BilinearUp2 up-samples input tensors by a factor of 2 using bi-linear interpolation;
BicubicResize4(input) increases the resolution of the input tensor to 4 times higher via bi-cubic up-sampling;
$Dense(\text{input}, \text{output\_size})$ is a densely-connected layer, which uses Xavier initialization for the kernel weights.

The architecture of our VSR generator G is:
\begin{center}
$\text{conc}(a_t,  W( g_{t-1}, v_t) )  \rightarrow  l_{in}$ ;
$C(l_{in}, 3, 64, 1), \text{ReLU}  \rightarrow  l_0$;\\
${ResidualBlock}(l_i)   \rightarrow  l_{i+1} \text{~with~} i = 0,...,n-1 $;\\
$CT(l_{n}, 3, 64, 2), \text{ReLU}  \rightarrow  l_{up2} $;
$CT(l_{up2}, 3, 64, 2), \text{ReLU}  \rightarrow  l_{up4}$;\\
$C(l_{up4}, 3, 3, 1), \text{ReLU}  \rightarrow  l_{res}$;
$\text{BicubicResize4}(a_t) + l_{res} \rightarrow  g_t \ .$\\
\end{center}
In \TecoGANsmall, there are 10 {sequential} residual blocks in the generator ( $l_{n} = l_{10} $ ), while the {TecoGAN generator}
has 16 residual blocks ( $l_{n} = l_{16} $ ).
Each ${ResidualBlock}(l_i)$ contains the following operations:
$C(l_{i}, 3, 64, 1)$, {ReLU} $\rightarrow r_{i}$; $ C(r_{i}, 3, 64, 1) + l_{i} \rightarrow l_{i+1} $.

The VSR $D_{s,t}$'s architecture is:
\begin{center}
${I}^{g}_{s,t} \; \text{or} \; {I}^{b}_{s,t} \rightarrow  l_{in} $;
$C(l_{in}, 3, 64, 1), \text{Leaky ReLU}  \rightarrow  l_0 $;\\
$C(l_0, 4, 64, 2), \text{BatchNorm}, \text{Leaky ReLU}  \rightarrow  l_1 $;
$C(l_1, 4, 64, 2), \text{BatchNorm}, \text{Leaky ReLU}  \rightarrow  l_2 $;\\
$C(l_2, 4, 128, 2), \text{BatchNorm}, \text{Leaky ReLU}  \rightarrow  l_3 $;
$C(l_3, 4, 256, 2), \text{BatchNorm}, \text{Leaky ReLU}  \rightarrow  l_4 $;\\
$Dense(l_4, 1), \text{sigmoid}  \rightarrow  l_{out} \ . $\\
\end{center}

VSR discriminators used in our variant models, DsDt, DsDtPP and DsOnly, have a the same architecture as
$D_{s,t}$. They only differ in terms of their inputs.

The flow estimation network F has the following architecture:
\begin{center}
$ \text{conc}(a_t,  a_{t-1})  \rightarrow  l_{in} $;
$ C(l_{in}, 3, 32, 1), \text{Leaky ReLU}  \rightarrow  l_0 $;\\
$ C(l_0, 3, 32, 1), \text{Leaky ReLU}, \text{MaxPooling}  \rightarrow  l_1 $;
$ C(l_1, 3, 64, 1), \text{Leaky ReLU}  \rightarrow  l_2 $;\\
$ C(l_2, 3, 64, 1), \text{Leaky ReLU}, \text{MaxPooling}  \rightarrow  l_3 $;
$ C(l_3, 3, 128, 1), \text{Leaky ReLU}  \rightarrow  l_4 $;\\
$ C(l_4, 3, 128, 1), \text{Leaky ReLU}, \text{MaxPooling}  \rightarrow  l_5 $;
$ C(l_5, 3, 256, 1), \text{Leaky ReLU}  \rightarrow  l_6 $;\\
$ C(l_6, 3, 256, 1), \text{Leaky ReLU}, \text{BilinearUp2}  \rightarrow  l_7 $;
$ C(l_7, 3, 128, 1), \text{Leaky ReLU}  \rightarrow  l_8 $;\\
$ C(l_8, 3, 128, 1), \text{Leaky ReLU}, \text{BilinearUp2}  \rightarrow  l_{9} $;
$ C( l_{9}, 3, 64, 1), \text{Leaky ReLU}  \rightarrow   l_{10} $;\\
$ C( l_{10}, 3, 64, 1), \text{Leaky ReLU}, \text{BilinearUp2}  \rightarrow l_{11} $;
$ C(l_{11}, 3, 32, 1), \text{Leaky ReLU}  \rightarrow  l_{12} $;\\
$ C(l_{12}, 3, 2, 1), \text{tanh}  \rightarrow  l_{out} $;
$ l_{out} * \text{MaxVel}  \rightarrow  v_{t} \ . $\\
\end{center}
Here, MaxVel is a constant vector, which scales the network output to the normal velocity range.

While F is the same for UVT tasks, UVT generators have an encoder-decoder structure:
{\small
\begin{center}
$\text{conc}(a_t,  W( g_{t-1}, v_t) )  \rightarrow  l_{in}$ ;
$C(l_{in}, 7, 32, 1),  \text{InstanceNorm}, \text{ReLU}  \rightarrow  l_0 $;\\
$C(l_{0}, 3, 64, 2),  \text{InstanceNorm}, \text{ReLU}  \rightarrow  l_1 $;
$C(l_{1}, 3, 128, 2),  \text{InstanceNorm}, \text{ReLU}  \rightarrow  l_2 $;\\
${ResidualBlock}(l_2+i) \rightarrow  l_{3+i} \text{~with~} i = 0,...,n-1 $;\\
$CT(l_{n+2}, 3, 64, 2),  \text{InstanceNorm}, \text{ReLU}  \rightarrow  l_{n+3}$;
$CT(l_{n+3}, 3, 32, 2),  \text{InstanceNorm}, \text{ReLU}  \rightarrow  l_{n+4}$;\\
$CT(l_{n+4}, 7, 3, 1), \text{tanh}  \rightarrow  l_{out} $\\
\end{center}
}
{\small ${ResidualBlock}(l_2+i)$}
contains the following operations:
$C(l_{2+i}, 3$, $128, 1)$,  {InstanceNorm}, {ReLU} $\rightarrow t_{2+i}$
;$C(t_{2+i}, 3, 128, 1)$, {InstanceNorm} $\rightarrow r_{2+i};$
$r_{2+i} + l_{2+i} \rightarrow l_{3+i} $. We use 10 residual blocks for all
UVT generators.

Since UVT generators are larger than the VSR generator, we also use a larger $D_{s,t}$ architecture:
{\small
\begin{center}
${I}^{g}_{s,t} \; \text{or} \; {I}^{b}_{s,t} \rightarrow  l_{in} $;
$C(l_{in}, 4, 64, 24), \text{ReLU}  \rightarrow  l_0 $;\\
$C(l_0, 4, 128, 2), \text{InstanceNorm}, \text{Leaky ReLU}  \rightarrow  l_1 $;
$C(l_1, 4, 256, 2), \text{InstanceNorm}, \text{Leaky ReLU}  \rightarrow  l_2 $;\\
$C(l_2, 4, 512, 2), \text{InstanceNorm}, \text{Leaky ReLU}  \rightarrow  l_3 $;
$Dense(l_3, 1) \rightarrow  l_{out} \ .$\\
\end{center}}
Again, all ablation studies use the same architecture and only differ
in terms of their inputs.

\section{Training Details}
\label{app:training}

We use the non-saturated GAN for VSR and LSGAN~\citep{mao2017least} for UVT
and both of them can prevent the gradient vanishing problem	of a standard  GAN~\citep{goodfellow2014generative}.
We employ a dynamic discriminator updating strategy,
i.e. discriminators are not updated when there is a large difference between $D({I}^{g}_{s,t})$ and $D({I}^{b}_{s,t})$.
While our training runs are generally very stable,
the training process could potentially be further improved with modern GAN algorithms, e.g. Wasserstein GAN~\citep{WGAN}.

To improve the stability of the adversarial training for the VSR task, we pre-train G and \fnet together with a simple $L^2$ loss of\linebreak
$\myavg \left \| g_t-b_t \right \|_{2} $ $ + \lambda_{w} \mathcal{L}_{warp}$ for 500k batches. Based on the pre-trained models, we use 900k batches for the proposed spatio-temporal adversarial training stage. Our training sequences has a length of 10 and a batch size of 4. A black image is used as the first previous frame of each video sequence. I.e., one batch contains 40 frames and with the \ppl{} loss formulation, the NN receives gradients from 76 frames in total for every training iteration.

In the pre-training stage of VSR, we train the F and a generator with 10 residual blocks.
An ADAM optimizer with $\beta = 0.9$ is used throughout.
The learning rate starts from $10^{-4}$ and decays by 50\mypercent{} every 50k batches until it reaches $2.5*10^{-5}$.
This pre-trained model is then used for all TecoGAN variants as initial state.
In the adversarial training stage of VSR, all TecoGAN variants are trained
with a fixed learning rate of $5*10^{-5}$.
The generators in DsOnly, DsDt, DsDtPP and \TecoGANsmall{} have 10 residual blocks,
whereas the TecoGAN model has 6 additional residual blocks in its generator.
Therefore, after loading 10 residual blocks from the pre-trained model,
these additional residual blocks are faded in smoothly with a factor of $2.5*10^{-5}$.
We found this growing training methodology~\citep{karras2017progressive},
to be stable and efficient in our tests. When training the VSR DsDt and DsDtPP,
extra parameters are used to balance the two cooperating discriminators properly.
Through experiments, we found $D_t$ to be stronger.
Therefore, we reduce the learning rate of $D_t$ to $1.5*10^{-5}$ in order to keep both discriminators balanced.
At the same time, a factor of 0.0003 is used on the temporal adversarial loss to the generator,
while the spatial adversarial loss has a factor of 0.001.
During the VSR training, input LR video frames are cropped to a size of $32\times32$.
In all VSR models, the Leaky ReLU operation uses a tangent of 0.2 for the negative half space.
Additional training parameters are listed in \myreftab{tab:training_details}.

{For the UVT task, a pre-training is not necessary for generators and} discriminators since temporal triplets are gradually faded in. Only a pre-trained \fnet model is reused.
Trained on specialized data-sets, we found UVT models to converge well with 100k batches of sequences in length of 6 and batch size of 1.

For all UVT tasks, we use a learning rate of $10^{-4}$ to train the first 90k batches and
the last 10k batches are trained with the learning rate decay from $10^{-4}$ to 0.
Images of the input domain are cropped into a size of $256\times256$ when training, the original size being $288\times288$.
The additional training parameters are also listed in \myreftab{tab:training_details}.
For UVT, $\mathcal{L}_\text{content}$ and $\mathcal{L}_{\phi}$ are only used to
improve the convergence of the training process.
We fade out $\mathcal{L}_\text{content}$ in the first 10k batches
and $\mathcal{L}_{\phi}$ is used for the first 80k and faded out in last 20k.

\begin{table}
\caption{Training parameters}
\label{tab:training_details}
{\setlength{\tabcolsep}{0.54pt}
\begin{tabular}{c|c|c|c|c|c}
\hline
VSR Param  & {DsOnly} &{DsDt}&{DsDtPP}&{\TecoGANsmall}& {TecoGAN}\\
\hline
{$\lambda_{a}$} & {1e-3} &
\multicolumn{2}{c|}{{Ds: 1e-3}, {Dt: 3e-4}} &
{1e-3} & {1e-3}\\
\hline
{$\lambda_{p}$} &  0.0 & 0.0 & \multicolumn{3}{c}{0.5} \\
\hline
{$\lambda_{\phi}$} &  \multicolumn{5}{c}{\small
0.2 for VGG and 1.0 for Discriminator} \\
\hline
{$\lambda_{\omega}$, $\lambda_{c}$} & \multicolumn{5}{c}{1.0, 1.0} \\
\hline
{ $^\text{learning-}_\text{rate}$ } &
{ 5e-5 } &
\multicolumn{2}{c|} { $^{\text{5e-5 for Ds. 1.5e-5 for Dt.}}_\text{5e-5 for G, F.}$} &
{ 5e-5 } &
{ 5e-5 }\\
\hline
\end{tabular}
}\\
{\setlength{\tabcolsep}{8pt}
\begin{tabular}{c|c|c|c|c}\hline
UVT Param  & {DsOnly} & {Dst} & {DsDtPP} & {TecoGAN}\\
\hline
{$\lambda_{a}$} & \multicolumn{2}{c|}{0.5} &
$^\text{Ds: 0.5}_\text{Dt: 0.3}$ & 0.5\\
\hline
{$\lambda_{p}$} &  0.0 & 0.0 & \multicolumn{2}{c}{100.0} \\
\hline
{$\lambda_{\phi}$} & \multicolumn{4}{c}{from $10^6$ decays to 0.0} \\
\hline
{$\lambda_{\omega}$} & \multicolumn{4}{c}{0.0, a pre-trained F is used for UST tasks} \\
\hline
{$\lambda_{c}$} & \multicolumn{4}{c}{10.0} \\
\hline

\end{tabular}
}
\end{table}

\section{Performance}
\label{app:performance}
TecoGAN is implemented in {\em TensorFlow}. While generator and discriminator are trained together, we only need
the trained generator network for the inference of new outputs after training,
i.e., the whole discriminator network can be discarded.
We evaluate the models on a Nvidia GeForce GTX 1080Ti GPU with 11G memory,
the resulting VSR performance for which is given in \myreftab{tab:metrics}.

The VSR \TecoGANsmall{} model and FRVSR have the same number of weights
(843587 in the generator network and 1.7M in F), and thus show
very similar performance characteristics with around 37 ms per frame.
The larger VSR TecoGAN model with 1286723 weights in the generator
is slightly slower than them, spending 42 ms per frame.
In the UVT task, generators spend around 60 ms per frame with a size of $512\times512$.
However, compared with models of DUF, EDVR and RBPN, which have 6 to 20 million weights,
the performance of TecoGAN is significantly better thanks to its reduced size.

\end{document}